\pdfoutput=1

\documentclass[11pt]{article}

\usepackage[final]{acl}

\usepackage{times}
\usepackage{latexsym}

\usepackage[T1]{fontenc}

\usepackage[utf8]{inputenc}

\usepackage{microtype}

\usepackage{inconsolata}

\usepackage{multirow}
\usepackage{graphicx}

\usepackage{arydshln}

\makeatletter

\newcommand{\Rmnum}[1]{\expandafter\@slowromancap\romannumeral #1@}
\makeatother

\usepackage{color}

\usepackage{booktabs}
\usepackage{CJKutf8}

\usepackage{quoting}
\quotingsetup{vskip=0pt}

\title{Human-LLM Hybrid Text Answer Aggregation for Crowd Annotations}

\author{Jiyi Li \\
  University of Yamanashi, Kofu, Japan \\
  \texttt{garfieldpigljy@gmail.com} }

\begin{document}
\maketitle
\begin{abstract}
The quality is a crucial issue for crowd annotations. Answer aggregation is an important type of solution. 
The aggregated answers estimated from multiple crowd answers to the same instance are the eventually collected annotations, rather than the individual crowd answers themselves.  
Recently, the capability of Large Language Models (LLMs) on data annotation tasks has attracted interest from researchers. Most of the existing studies mainly focus on the average performance of individual crowd workers; several recent works studied the scenarios of aggregation on categorical labels and LLMs used as label creators. However, the scenario of aggregation on text answers and the role of LLMs as aggregators are not yet well-studied. 
In this paper, we investigate the capability of LLMs as aggregators in the scenario of close-ended crowd text answer aggregation. 
We propose a human-LLM hybrid text answer aggregation method with a Creator-Aggregator Multi-Stage (CAMS) crowdsourcing framework. We make the experiments based on public crowdsourcing datasets. The results show the effectiveness of our approach based on the collaboration of crowd workers and LLMs. 

\end{abstract}

\section{Introduction}
Because of the ability or diligence of the workers, the quality of crowdsourced annotations is a crucial issue. One solution of quality control is collecting redundant annotations by assigning multiple crowd workers to one instance and aggregating the multiple answers to this instance into an estimated answer which has high reliability. 

There are many existing works in the crowdsourcing area on answer aggregation for various types of annotations. 
For example, for the classification tasks, crowd workers provide categorical labels to the instances \cite{mj,gladlong,cbcclong}. 
For some ranking tasks, workers provide pairwise preference labels, i.e., whether an instance is preferred to the other instance \cite{crowdbtlong,ChenIntransitivityWSDMlong,HBTL}. 
In this paper, we mainly focus on the tasks with text answers, which can be the translations of given sentences \cite{textaggregationdataset}, free-text rationales for explaining NLP models \cite{ExplainNLPSurvey,ExplainNLPData}, and so on. We mainly consider the close-ended text answer annotations, i.e., the target answers are determined to some extent and can be considered as the golden standards, while the aggregation methods estimate these golden standards based on multiple crowd answers. 
Table \ref{tab:example} shows examples of categorical label aggregation and text answer aggregation. We denote these answer aggregation methods as model aggregators.

\begin{table}[!t]
\scriptsize
\setlength\tabcolsep{1.5pt}
\caption{\label{tab:example} Examples of aggregation tasks for crowd annotations. $a_i^j$ represents the crowd answer to an instance $q_i$ by a crowd worker $w_j$. $\hat{z}_i$ represents the estimated answer by an aggregation method. To aid in understanding the crowd text answer aggregation task in this study, we also provide an example of the crowd categorical label aggregation task in other studies.} 
\hfill
\vspace{0.1cm}
\begin{minipage}
{0.01\textwidth}
$q_1$
\end{minipage}
\begin{minipage}{0.1\textwidth}
\centering
\includegraphics[height=1.2cm]{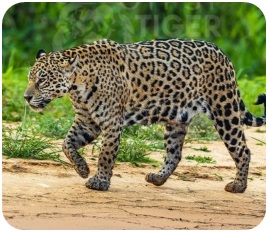}
\end{minipage}
\begin{minipage}{0.35\textwidth}
\centering
\begin{tabular}{cccccc|c}
\multicolumn{6}{c}{\Rmnum{1}: Categorical Answer (Label) Aggregation Task} \\ 
\hline 
$a_1^1$ & $a_1^2$ & $a_1^3$ & $a_1^4$ & $a_1^5$ & $a_1^j$ & $\hat{z}_1$ \\ 
leopard & jaguar & cheetah & leopard & jaguar & ... & leopard \\
\hline
\end{tabular}
\end{minipage}
\begin{minipage}{0.49\textwidth}
\centering 
\begin{tabular}{c|cc|c}
\multicolumn{3}{c}{\Rmnum{2}: Text Answer Aggregation Task (this study focuses this task)} \\
\hline 

\multirow{5}{*}{\parbox{1.2cm}{$q_1$:\begin{CJK*}{UTF8}{goth}
私には車椅子なんか全く必要が無い
\end{CJK*} (Translate into English)}} & $a_1^1$: & \parbox{4.6cm}{For me it is not necessary to have a wheelchair at all.} & \multirow{5}{*}{\parbox{1.2cm}{$\hat{z}_1$: I don't need a wheelchair at all.}} \\\cline{2-3}
 & $a_1^2$: & \parbox{4.6cm}{A wheelchair is not necessary for me at all.} & \\\cline{2-3}
 & $a_1^3$: & \parbox{4.6cm}{The wheelchair is totally unnecessary for me.} & \\\cline{2-3}
 & $a_1^4$: & \parbox{4.6cm}{There is no need for me to have such a wheelchair.} & \\\cline{2-3}
 & $a_1^5$: & \parbox{4.6cm}{I never need wheelchair.} & \\\cline{2-3}
 & $a_1^j$: & \parbox{4.6cm}{......} & \\
\hline
\end{tabular}
\end{minipage}
\end{table}

Recently, the capability of Large Language Models (LLMs) on data annotation tasks has attracted interest from researchers.~Because LLMs are cheaper than crowd workers for annotating the instances (e.g., \citet{CrowdChatgptTextAnnotation} reported that ChatGPT is about twenty times cheaper than MTurk in their experiments), whether LLMs can outperform crowdsourcing is one of the issues concerned. 
\citet{veselovsky2023artificial} found that the crowd workers on MTurk have been recently using LLMs to complete the crowdsourcing tasks. 
Some works verified this issue with the average performance of individual crowd workers and LLMs on some specific tasks by collecting new datasets for their target tasks \cite{CrowdChatgptSocialComputing,CrowdChatgptTextAnnotation,CrowdChatgptTwitter, CrowdChatgptIntentClassification,annollm}. 
These LLM studies mainly concentrate on the average performance of individual crowd workers. 
However, the aggregated answers are the eventually collected annotations, rather than the crowd answers themselves. Therefore, the scenarios involving crowd answer aggregation need further study. 

Several recent works \cite{llmcrowdaggICASSP2024,llmcrowdaggCHI24} studied the scenario of answer aggregation on the crowd categorical labels in the classification tasks and LLMs are used as creators of the labels. 
The scenario of aggregation on other types of crowd annotations such as text answers and other types of LLM roles such as aggregators are not yet well-studied.
We focus on investigating the capability of LLMs as aggregators on crowd text answer aggregation. 
LLMs can naturally utilize multiple text answers as the input, aggregate them, and generate the estimated answer. In other words, we can utilize a LLM as a LLM aggregator. 

In crowdsourcing area, because single-stage frameworks only collect all annotations by crowd workers at once, which is limited for the requesters to flexibly set diverse additional mechanisms for improving the data quality, multi-stage frameworks are proposed, e.g., Creation-Review two stages \cite{twostagequality} for image description and logo design, Find-Fix-Verify three stages \cite{findfixverify} for shortening long texts, and so on. \citet{llmcrowdpipeline} utilized LLMs in some multi-stage crowdsourcing frameworks. 

In this paper, we propose a human-LLM hybrid text answer aggregation method based on a Creator-Aggregator Multi-Stage (CAMS) crowdsourcing framework.  
In the first stage, crowd workers (Crowd Creators) provide the raw text answers. In the next stage, we ask crowd workers (Crowd Aggregators) and LLMs (LLM Aggregators) to aggregate the raw crowd answers to each instance. In the final stage, we utilize the model aggregators on the combinations of three resources of answers with the workers, including raw crowd answers, the estimated answers by crowd aggregators, and the estimated answers by LLM aggregators, to estimate the true answers. We construct experiments with public crowdsourcing datasets. 
The contributions are as follows. 
\begin{itemize}
\item We investigate the capability of LLMs as aggregators in the scenario of crowd text answer aggregation for the quality control of crowdsourced close-ended text answer annotations tasks. 
\item We propose a human-LLM hybrid text answer aggregation method with a Creator-Aggregator Multi-Stage (CAMS) crowdsourcing framework which is a solution based on the collaboration of crowd workers and LLMs. 
\item We conducted experiments using public crowdsourcing datasets for crowd creators and additional data for crowd aggregators. The results show that both crowd aggregators and LLM aggregators can produce higher-quality estimated answers compared to raw crowd answers from crowd creators. Our CAMS approach with model aggregators can further improve the answer quality based on the combinations of three resources of workers and answers. 
\end{itemize}

\section{Related Work}
There are many existing works in the crowdsourcing area on answer aggregation for various types of annotations. 
For the categorical labels, some approaches jointly estimate worker ability and true answers using 
the maximum entropy principle~\cite{Minimax_Entropy}, 
and Bayesian inference~\cite{cbcclong}. 
More sophisticated models incorporate task difficulty~\cite{gladlong} 
and their Bayesian treatments~\cite{wauthier2011bayesian,DARE}. 
There are also other methods based on diverse techniques such as autoencoders \cite{LAA}, graph mining \cite{graphexpert}, and so on \cite{IterativeReduction,workersimilarity,crowdpr}. Besides answer aggregation approaches, There is another type of approach that directly trains the classification models with the noisy crowd labels \cite{lfcxfeature}. In addition, some approaches consider reducing the budget while preserving the utility of the aggregated labels \cite{budgetcost}.  

For pairwise preference comparison labels ~\cite{preferencesurvey}, 
a typical solution is the Bradley-Terry model and its various extensions or generalizations to diverse settings  \cite{2dbt,ChenIntransitivityWSDMlong,ChenIntransitivityKDDlong,crowdbtlong,peergrading,rankingclusteringlong,HBTL,affectipartialrank,contextpairwisecrowdopinions,moepra}. There are also other types of methods, e.g., matrix completion~\cite{matrix1long,matrix2long}. 

There are also methods for pairwise similarity labels \cite{sne,tsnelong,crowdclustering,semicrowdclustering}, triplet similarity labels \cite{stelong,crowdtriplet,multiviewcrowdtriplet}, numerical data \cite{catd} and text data \cite{textaggregationdataset,textaggregation}. 
We share our crowdsourcing datasets, including various data types, which can be utilized for research on topics such as answer aggregation for crowd annotations at
\url{https://github.com/garfieldpigljy/ljycrowd}. In this paper, we mainly consider the close-ended text answer annotations. 

In crowdsourcing area, because single-stage frameworks only collect all annotations by crowd workers at once, which is limited for the requesters to flexibly set diverse additional mechanisms for improving the data quality, multi-stage frameworks are proposed, e.g., Partition-Map-Reduce \cite{PartitionMapReduce} for writing essays, Price-Divide-Solve \cite{pricedividesolve} for writing essays, Creation-Review \cite{twostagequality} for image description and logo design, Find-Fix-Verify \cite{findfixverify} for shortening long texts,  Pair-Compare-Sort \cite{PairCompareSort} for sorting, and so on. \citet{llmcrowdpipeline} utilized LLMs in some multi-stage crowdsourcing frameworks, e.g., the Find-Fix-Verify framework. Furthermore, \citet{aichainVScrowdpipeline} compared the similarities and differences between LLM chains and multi-stage crowdsourcing frameworks. In this paper, we propose a specialized creator-aggregator multi-stage crowdsourcing framework for crowd text answer aggregation tasks.

\section{Preliminary}\label{sec:preliminary}
\begin{figure*}[!t]
\centering
\includegraphics[width=11cm]{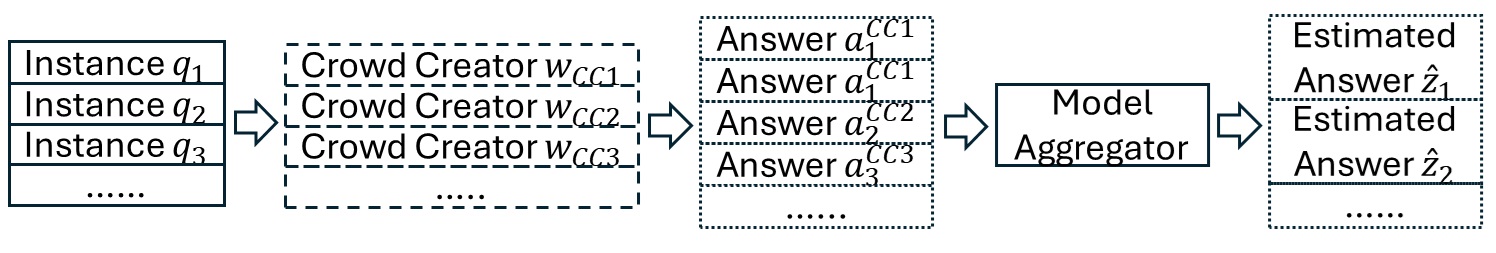}
\caption{Existing Single-Stage framework for crowdsourced text answer aggregation.}
\label{fig:singlestageframework} 
\end{figure*}

\begin{figure*}[!t]
\centering
\includegraphics[width=16cm]{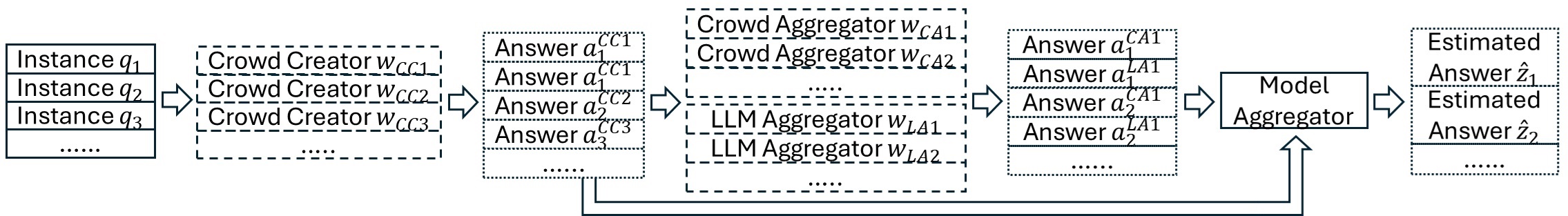}
\caption{Our Creator-Aggregator Multi-Stage (CAMS) framework for crowdsourced text answer aggregation. }
\label{fig:multistageframework} 
\end{figure*}

\subsection{Problem Setting of Answer Aggregation}
We formulate the general problem setting of crowd text answer aggregation for quality control in close-ended text answer annotations. 
We define the instance set $\mathcal{Q}=\{q_i\}_i$, worker set $\mathcal{W}=\{w_j\}_j$, answer set $\mathcal{A}=\{a_i^j\}_{i,j}$, where $a_i^j$ represents the answer to question $q_i$ by worker $w_j$, true answer set $\mathcal{Z}=\{z_i\}_i$ and estimated answer set $\hat{\mathcal{Z}}=\{\hat{z}_i\}_i$. 
The answer set of an instance $q_i$ is $\mathcal{A}_i$; the answer set of a worker $w_j$ is $\mathcal{A}^j$. Because the number of instances can be large, a worker is not necessary to annotate all instances. 

In the setting of the crowd answer aggregation, ground truths cannot be used by model aggregators because the purpose of data annotation tasks is collecting and estimating the ground truth.  
Therefore, abstractive aggregation methods, such as abstractive text summarization, which require ground truth for training, are not easily adaptable for use as model aggregators. 
Existing model aggregators for crowd text answer aggregation (e.g., \cite{ModelEnsembleSummary, textaggregationdataset}) are mainly extractive aggregation methods and select the potential optimal one from multiple crowd answers. 
Given the instance set $\mathcal{Q}$, worker set $\mathcal{W}$ and answer set $\mathcal{A}$, a model aggregator estimates the true answer $\hat{\mathcal{Z}}$ by selecting one optimal answer $\hat{z}_i$ from $\mathcal{A}_i$ for each instance $q_i$, i.e., the answer aggregation can be defined as $\hat{\mathcal{Z}} = \textnormal{Agg}(\mathcal{Q}, \mathcal{W}, \mathcal{A})$. Note that, in this study, only model aggregators are extractive aggregators. Crowd aggregators and LLM aggregators are abstractive aggregators because they read the raw crowd answers and generate aggregated answers in their own words. 

\subsection{Model Aggregators}
We utilize several aggregation approaches from existing works as the model aggregators. 
These approaches utilize an encoder $e(\cdot)$ to convert the text into embeddings. 

\noindent
\textbf{Sequence Majority Voting (SMV): } 
It is a na\"ive but typical answer aggregation approach that is adapted from majority voting. 
For each instance, it estimates the embeddings of the true answers by $\hat{e}_i = mean(e(\mathcal{A}_i))$, and selects the worker answer $\hat{z}_i = \arg\max_{a_i^j} sim(e(a_i^j),\hat{e}_i).$ 

\noindent
\textbf{Sequence Maximum Similarity (SMS): } For each instance, SMS \cite{ModelEnsembleSummary} selects the worker answer which has the largest sum of similarity with other answers to this question. It can be regarded as creating a kernel density estimator and selecting the maximum density answer. The kernel function uses the cosine similarity. The formulation is  
$\hat{z}_i = \arg\max_{a_i^{j_1}}\sum_{j_1\neq j_2} sim(e(a_i^{j_1}),e(a_i^{j_2})).$ 

\noindent
\textbf{Reliability Aware Sequence Aggregation (RASA): } 
RASA \cite{textaggregationdataset} models the worker reliability $\theta$ to strengthen the influences of answers provided by the workers with higher estimated reliability. 
It first estimates the embeddings of the true answers considering worker reliability. It  iteratively estimates $\theta$ and $\hat{e}_i$ until convergence, $\theta_j = \frac{{\chi}^2_{(\alpha/2,|\mathcal{A}^j|)}}{\sum_i||e(a_i^j)-\hat{e}_i||^2}, \hat{e}_i = \frac{\sum_j \theta_je(a_i^j) }{\sum_j \theta_j},$ 
where ${\chi}^2$ is the chi-squared distribution and the significance level $\alpha$ is set as 0.05 empirically. $\hat{e}_i$ can be initialized by using the SMV method. 
After that, it selects a worker answer that is most similar to the embeddings of the estimated true answer $\hat{e}_i$. 

SMV and SMS are instance-wise aggregators that utilize the answers of an instance as the input. RASA is a dataset-wise aggregator that leverages the potential relations among workers and instances to estimate the answers for all instances. From this viewpoint, the crowd aggregators and LLM aggregators are instance-wise aggregators.

\section{Human-LLM Hybrid Aggregation}
\begin{table}[t!]
\small
\caption{\label{tab:dataset_statistics} Statistics of the datasets, $|\cdot|$ is the size of the sets. The answers $\mathcal{A_{C.C.}}$ by crowd creators $\mathcal{W_{C.C.}}$ is from CrowdWSA; the answers $\mathcal{A_{C.A.}}$ by crowd aggregators $\mathcal{W_{C.A.}}$ are collected by this work. }
\begin{center}
\begin{tabular}{c|ccccc}
\hline
Data & $|\mathcal{Q}|$ & $|\mathcal{W_{C.C.}}|$ &  $|\mathcal{A_{C.C.}}|$ & $|\mathcal{W_{C.A.}}|$ &  $|\mathcal{A_{C.A.}}|$ \\ \hline
J1 & 250 & 70 & 2,490 & 106 & 1250 \\ 
T1 & 100 & 42 & 1,000 & 69 & 500 \\ 
T2 & 100 & 43 & 1,000 & 71 & 500 \\  
\hline
\end{tabular}
\end{center}
\end{table}

Most of the existing work on crowdsourced annotation is based on the single-stage framework, i.e., all crowd answers are collected from crowd creators at once and the model aggregator estimates the true answers on these raw crowd answers directly. 
Figure \ref{fig:singlestageframework} shows an example of a single-stage framework for crowdsourced text answer annotation. Single-stage frameworks are limited for the requesters to flexibly set diverse additional mechanisms for improving the data quality. Multi-stage frameworks are candidate solutions to improve flexibility and quality. 

Because the crowd answers in this study are text, it is intuitive to ask LLMs to read the multiple crowd text answers to a given instance, and generate the estimated text answer. LLMs can be used as LLM aggregators. 
Similarly, crowd workers can be asked to provide the estimated answer by reading the raw crowd answers of an instance, and be used as crowd aggregators. We regard the Crowd Creator (C.C.), Crowd Aggregator (C.A.), and LLM Aggregator (L.A.) as three different types of workers. 
To distinguish these three resources of workers and answers, we define crowd creator workers and their answers as $\mathcal{W}_{C.C.}$ and $\mathcal{A}_{C.C.}$, crowd aggregator workers and their answers as $\mathcal{W}_{C.A.}$ and $\mathcal{A}_{C.A.}$, and LLM aggregator workers and their answers as $\mathcal{W}_{L.A.}$ and $\mathcal{A}_{L.A.}$.
Note that crowd creators $\mathcal{W}_{C.C.}$ and crowd aggregators $\mathcal{W}_{C.A.}$ are allowed to have overlaps; they are not necessary to be the same or completely different. This is a tractable setting for requesters in practice, because $\mathcal{A}_{C.C.}$ and $\mathcal{A}_{C.A.}$ are collected in separate stages. 

To effectively leverage these three resources of workers and answers in one crowdsourcing framework, we propose a Creator-Aggregator Multi-Stage (CAMS) framework. Figure \ref{fig:multistageframework} shows the proposed CAMS framework. In the first stage, crowd creators provide the raw crowd answers. In the next stage, we ask crowd aggregators and/or LLM aggregators to aggregate the raw crowd answers into estimated answers in their own words. In the final stage, we utilize the model aggregators on the combinations of three resources of workers and answers to generate the estimated true answers. 

There are different versions of our approach. We can combine two or three resources of workers and answers to create the merged sets, which are then used as input for the model aggregator to estimate the true answers. We provide an example based on a version that utilizes all three resources of workers and answers. Given the instance set $\mathcal{Q}$, we can construct the worker set $\mathcal{W}$ by merging three worker sets with $\mathcal{W}$ = $\mathcal{W}_{C.C.}$+$\mathcal{W}_{C.A.}$+$\mathcal{W}_{L.A.}$, and the answer set $\mathcal{A}$ by merging three answer sets with $\mathcal{A}$ = $\mathcal{A}_{C.C.}$+$\mathcal{A}_{C.A.}$+$\mathcal{A}_{L.A.}$. We use the symbol `+' rather than the symbol `$\cup$' for $\mathcal{W}$ and $\mathcal{A}$. Because the answer creation and answer aggregation are different tasks, if the same crowd worker provides answers as both crowd creator and crowd aggregator, he/she may have different behaviors and reliabilities in these two roles. We thus treat such a worker as different workers when merging the worker and answer resources. 
We can then utilize the model aggregators on these $\mathcal{Q}$, $\mathcal{W}$, and $\mathcal{A}$ to estimate the true answers, i.e., $\hat{\mathcal{Z}} = \textnormal{Agg}(\mathcal{Q}, \mathcal{W}, \mathcal{A})$ where $\textnormal{Agg}$ is one of SMV, SMS or RASA. 

\begin{table*}[t!]
\small
\setlength\tabcolsep{1.5pt}
\caption{\label{tab:singleworkerresults} Quality of individual answers by Crowd Creators (C.C.), Crowd Aggregators (C.A.) and LLM Aggregators (L.A., GPT-4 (O)) on GLEU, METEOR, and embedding similarity. }
\begin{center}
(a) GLEU\\
\begin{tabular}{c|ccc|ccc|ccc|ccc|ccc}
\hline
\multirow{3}{*}{Data} & \multicolumn{3}{c|}{MIN} & \multicolumn{3}{c|}{MEAN} & \multicolumn{3}{c|}{MAX} & \multicolumn{3}{c|}{STD} & \multicolumn{3}{c}{TIAA} \\ 
 & C.C. & C.A. & L.A.(O) & C.C. & C.A. & L.A.(O) & C.C. & C.A. & L.A.(O) & C.C. & C.A. & L.A.(O) & C.C. & C.A. & L.A.(O) \\ 
\hline
J1 & 0.0724 & 0.0000 & \textbf{0.2706} & 0.1868 & 0.2124 & \textbf{0.2729} & 0.5948 & \textbf{1.0000} & 0.2756 & 0.0915 & 0.1399 & 0.0018 & 0.1798 & 0.2746 & 0.7438 \\
T1 & 0.0669 & 0.0217 & \textbf{0.2144} & 0.1764 & 0.1874 & \textbf{0.2184} & 0.5534 & \textbf{0.7895} & 0.2283 & 0.0818 & 0.1297 & 0.0051 & 0.2533 & 0.3473 & 0.7809 \\
T2 & 0.0503 & 0.0000 & \textbf{0.2025} & 0.1716 & 0.1817 & \textbf{0.2136} & 0.4540 & \textbf{0.6842} & 0.2244 & 0.0838 & 0.1190 & 0.0083 & 0.2377 & 0.3371 & 0.7589 \\
\hline
\end{tabular}
\vspace{0.1cm}\\
(b) METEOR\\
\begin{tabular}{c|ccc|ccc|ccc|ccc|ccc}
\hline
\multirow{2}{*}{Data} & \multicolumn{3}{c|}{MIN} & \multicolumn{3}{c|}{MEAN} & \multicolumn{3}{c|}{MAX} & \multicolumn{3}{c|}{STD} & \multicolumn{3}{c}{TIAA}\\ 
 & C.C. & C.A. & L.A.(O) & C.C. & C.A. & L.A.(O) & C.C. & C.A. & L.A.(O) & C.C. & C.A. & L.A.(O) & C.C. & C.A. & L.A.(O) \\ 
\hline
J1 & 0.2041 & 0.0490 & \textbf{0.4987} & 0.3762 & 0.4156 & \textbf{0.5049} & 0.7464 & \textbf{0.9985} & 0.5103 & 0.1037 & 0.1569 & 0.0039 & 0.3791 & 0.4910 & 0.8683 \\
T1 & 0.1781 & 0.0394 & \textbf{0.4247} & 0.3771 & 0.3909 & \textbf{0.4294} & 0.7163 & \textbf{0.9170} & 0.4392 & 0.0987 & 0.1666 & 0.0051 & 0.4763 & 0.5670 & 0.8918 \\
T2 & 0.1523 & 0.0476 & \textbf{0.4102} & 0.3630 & 0.3840 & \textbf{0.4278} & 0.6473 & \textbf{0.8288} & 0.4407 & 0.1010 & 0.1619 & 0.0103 & 0.4402 & 0.5492 & 0.8696 \\
\hline
\end{tabular}
\vspace{0.1cm}\\
(c) Embedding Similarity\\
\begin{tabular}{c|ccc|ccc|ccc|ccc|ccc}
\hline
\multirow{2}{*}{Data} & \multicolumn{3}{c|}{MIN} & \multicolumn{3}{c|}{MEAN} & \multicolumn{3}{c|}{MAX} & \multicolumn{3}{c|}{STD} & \multicolumn{3}{c}{TIAA} \\ 
 & C.C. & C.A. & L.A.(O) & C.C. & C.A. & L.A.(O) & C.C. & C.A. & L.A.(O) & C.C. & C.A. & L.A.(O) & C.C. & C.A. & L.A.(O) \\ 
\hline
J1 & 0.4233 & 0.2243 & \textbf{0.7216} & 0.6620 & 0.6857 & \textbf{0.7313} & 0.8795 & \textbf{1.0000} & 0.7379 & 0.0687 & 0.1135 & 0.0053 & 0.6583 & 0.7286 & 0.9322 \\
T1 & 0.5827 & 0.1282 & \textbf{0.7758} & 0.7260 & 0.7336 & \textbf{0.7786} & 0.9043 & \textbf{1.0000} & 0.7824 & 0.0687 & 0.1511 & 0.0029 & 0.7252 & 0.7850 & 0.9532 \\
T2 & 0.4726 & 0.2700 & \textbf{0.7602} & 0.7077 & 0.7215 & \textbf{0.7651} & 0.8958 & \textbf{1.0000} & 0.7679 & 0.0861 & 0.1250 & 0.0031 & 0.7060 & 0.7752 & 0.9343 \\
\hline
\end{tabular}
\end{center}
\end{table*}

\section{Experiments}
\subsection{Datasets}
To investigate our study, we need datasets that contain multiple crowd text answers for each instance as well as the golden answers. We thus utilize the real datasets J1, T1, and T2 in CrowdWSA \cite{textaggregationdataset} which involve the translated sentences of the target language by multiple crowd workers for the sentences of the source language as well as the golden translations. Table \ref{tab:dataset_statistics} lists the statistics of the three datasets. 
The crowd workers in these datasets are regarded as the crowd creators (C.C.) in our CAMS framework. Each crowd creator provides answers for a subset of all instances. 

\subsection{Experimental Settings}
\subsubsection{Crowd Aggregators}
For crowd aggregators (C.A.), we utilize a commercial crowdsourcing platform Lancers\footnote{https://www.lancers.co.jp/} to collect the estimated answers. Each instance is assigned to five crowd aggregators. Each crowd aggregator provides estimated answers for a subset of all instances. The following instructions and task descriptions are shown to the crowd aggregators when deploying the micro crowdsourcing tasks: 

\noindent
{\quoting[leftmargin=0.25cm,rightmargin=0cm]
Instructions: \textit{The data is collected for use in academic research. The collected data will be used for research on natural language processing and machine learning. The workers' nicknames will not be disclosed to any third party other than the person requesting this task. }
\endquoting}

\noindent
{\quoting[leftmargin=0.25cm,rightmargin=0cm]
Descriptions: \textit{The ten English short sentences below were written by ten non-native English speakers who translated the same one Japanese short sentence from Japanese. Read each of the English short sentences below, infer the meaning of the original Japanese text, and based on your understanding, create what you consider to be one appropriate short sentence each in Japanese and English. }
\endquoting}

The original text instructions and task descriptions used at Lancers are in Japanese. Table \ref{tab:dataset_statistics} lists the statistics of the answers $\mathcal{A_{C.A.}}$ by crowd aggregators $\mathcal{W_{C.A.}}$ collected by this work. 

\subsubsection{LLM Aggregators}
For LLM aggregators (L.A.), we utilize GPT-4\footnote{The version is gpt-4-0125-preview.} and Gemini\footnote{The version is gemini-1.5-pro-latest in May 9-14th 2024.}. L.A.(O) uses GPT-4 and L.A.(G) uses Gemini. Because Gemini results are almost always worse than GPT-4 results, we only put GPT-4 results in the main manuscript and put Gemini results in the Appendix Section. We utilize different temperature values to construct multiple LLM aggregators. The main results are based on five LLM aggregators with temperatures [0,0.25,0.5,0.75,1], with one trial conducted for each temperature. Each LLM aggregator provides answers for all instances. The following prompt is presented to the LLMs: 

\noindent
{\quoting[leftmargin=0.25cm,rightmargin=0cm]
Prompt: \textit{The following ten short English sentences were written by ten non-native English speakers who translated the same Japanese short text from Japanese to English. Please read each English sentence and infer the meaning of the original Japanese text, and based on your understanding, provide the appropriate Japanese original text and its English translation. Only output the text without showing "Japanese original text" and "English translation". \textbackslash n\textbackslash n
<1. sentence 1>\textbackslash n<2. sentence 2>\textbackslash n......<10. sentence 10>\textbackslash n.
Output format separated by " and <tab>: "Japanese original text. " <tab> "its English translation. 
 " \textbackslash n\textbackslash n} 
\endquoting}

Regarding the budget cost associated with crowd aggregators and LLM aggregators, LLMs are generally cheaper than crowd workers for annotating instances. For example, \citet{CrowdChatgptTextAnnotation} reported that ChatGPT is about twenty times cheaper than MTurk in their experiments. 
In our experiments, LLM aggregators cost less than 0.01 dollars per instance, while crowd aggregators cost around 0.36 dollars per instance, indicating a cost difference of more than thirtyfold. 

\subsubsection{Model Aggregators}
For the model aggregators, following CrowdWSA, we use the answer aggregation approaches SMV, SMS, and RASA, and utilize universal sentence encoder\footnote{https://tfhub.dev/google/universal-sentence-encoder/4. This version is newer than the one used in the public implementation of CrowdWSA, thus the results of model aggregators on the answers of only crowd creators are different from those reported in CrowdWSA. }\cite{USE} as the text encoder. 

\subsubsection{Evaluation Metrics}
We also use three evaluation metrics to verify the proposal. Two of them are GLEU\footnote{We use NLTK package: nltk.translate.gleu\_score} and METEOR\footnote{We use NLTK package: nltk.translate.meteor\_score}. We compute the average scores between estimated answers and true answers.  
The other one is the average cosine similarity between the encoder embeddings of the estimated answers and the true answers. We utilize the original target sentences in the Japanese-English parallel corpus used by CrowdWSA as the ground truth. 

\subsection{Q1: Quality of Individual Answers by Crowd workers and LLMs}\label{sec:experiments_individual}
We investigate the individual performance of crowd creators, crowd aggregators, and LLM aggregators from the viewpoint of the quality of answers they provide or generate. Table \ref{tab:singleworkerresults} lists the results. 
First, the results in the columns of "Mean" show that both crowd aggregators and LLM aggregators can generate estimated answers with higher quality than raw crowd answers from crowd creators. 

Second, the results in the columns of "MIN", "MEAN", and "MAX" show that LLMs outperform bad and average crowd workers, but are worse than good crowd workers. Considering that model aggregators can amplify the influence of good workers when estimating true answers, a human-LLM hybrid aggregation with merged answer sets and model aggregators in the final stage has the potential to achieve better performance than aggregation results only based on crowd workers or LLMs. 

Furthermore, an observation is that the standard deviations ("STD" column) of LLM aggregators are much smaller than those of crowd creators and crowd aggregators. In such close-ended text answers, multiple LLM aggregators constructed using the same LLM with various temperature values (including relatively high values) have lower diversity than crowd creators and crowd aggregators. A potential disadvantage of low diversity is if the answer to an instance is not good, the LLM may not generate a good answer by re-generation or using diverse temperature values. High diversity in the answers is not bad for answer aggregation, because it provides a broader exploration of candidate answers and aggregation methods are proposed to estimate good workers and answers from raw answers. 

We also investigate the Inter-Annotator Agreement (IAA) among different types of workers. Because Cohen's Kappa for IAA is a statistical measure used to quantify the level of agreement between two raters who classify items into categories, while our annotations are text answers, we define a Text-Inter-Annotator Agreement (TIAA) for text answers by computing the scores between two answers to the same instance, $\kappa_{text} = \frac{1}{|\mathcal{Q}|}\sum_i \frac{1}{\mathcal{N}_{j_1,j_2}}\sum_{j_1,j_2} \mathcal{G}(a_i^{j_1},a_i^{j_2})$. $\mathcal{G}$ can be GLEU, METEOR, and Embedding Similarity. Higher TIAA means lower diversity in the answers. The results in the "TIAA" columns are based on GLEU and also show the lower diversity of LLM aggregators than crowd creators and crowd aggregators. We address this issue and candidate solution in the limitation section. 

\begin{table}[!t]
\small
\setlength\tabcolsep{3pt}
\caption{\label{tab:aggregatorperformance} Performance of different aggregators, C.A., L.A. (GPT-4 (O)), and Model Aggregator on the raw crowd answers $\mathcal{A_{C.C.}}$. }
\begin{center}
\begin{tabular}{c|cccccc}
\hline
\multirow{2}{*}{Data} & \multirow{2}{*}{C.A.} & \multicolumn{1}{c}{L.A.} & \multicolumn{3}{c}{Model Aggregator} \\
&  & (O) & SMV & SMS & RASA \\ 
\hline 
\multicolumn{6}{c}{(a) GLEU}
\\
J1 & 0.2124 & \textbf{0.2729} & 0.1930 & 0.2489 & 0.2537 \\ 
T1 & 0.1874 & 0.2184 & 0.1740 & 0.2310 & \textbf{0.2376} \\ 
T2 & 0.1817 & 0.2136 & 0.1616 & 0.2189 & \textbf{0.2340} \\ 
\hline
\multicolumn{6}{c}{(b) METEOR}
\\
J1 & 0.4156 & \textbf{0.5049} & 0.3861 & 0.4666 & 0.4745 \\  
T1 & 0.3909 & 0.4294 & 0.3786 & 0.4548 & \textbf{0.4718} \\  
T2 & 0.3840 & 0.4278 & 0.3604 & 0.4426 & \textbf{0.4653} \\ 
\hline
\multicolumn{6}{c}{(c) Embedding Similarity}
\\
J1 & 0.6857 & 0.7313 & 0.6732 & \textbf{0.7426} & 0.7414 \\ 
T1 & 0.7336 & 0.7786 & 0.7245 & 0.7935 & \textbf{0.8020} \\  
T2 & 0.7215 & 0.7651 & 0.7105 & 0.7860 & \textbf{0.7935} \\ 
\hline
\end{tabular}
\end{center}
\end{table}

\begin{table}[!t]
\scriptsize
\caption{\label{tab:aggresults_gleu} Results of text answer aggregation: GLEU; LLM aggregator (GPT-4 (O)).}
\begin{center}
\begin{tabular}{cccc|c|cc}
\hline
\multicolumn{4}{c|}{Answers} & \multicolumn{3}{c}{Model Aggregator} \\
 & $\mathcal{A}_{C.C.}$ & $\mathcal{A}_{C.A.}$ & $\mathcal{A}_{L.A.}$ & SMV & SMS & RASA \\ 
\hline
\multicolumn{7}{c}{J1}
\\
\multirow{1}{*}{\Rmnum{1}} & $\bigcirc$ & & & 0.1930 & 0.2489 & 0.2537 \\
\cdashline{1-7}
\multirow{2}{*}{\Rmnum{2}} & & $\bigcirc$ & & 0.2260 & 0.2740 & 0.2444 \\
 & & & $\bigcirc$(O) & \textbf{0.2729} & 0.2770 & 0.2740 \\
\cdashline{1-7}
\multirow{1}{*}{\Rmnum{3}} & $\bigcirc$ & $\bigcirc$ & & 0.2040 & 0.2873 & 0.2708 \\
\cdashline{1-7}
\multirow{3}{*}{\Rmnum{4}} & $\bigcirc$ & & $\bigcirc$(O) & 0.2196 & 0.2846 & 0.2742 \\
 & & $\bigcirc$ & $\bigcirc$(O) & 0.2494 & 0.2962 & 0.2732 \\
 & $\bigcirc$ & $\bigcirc$ & $\bigcirc$(O) & 0.2212 & \textbf{0.3003} & \textbf{0.2812} \\
\hline
\multicolumn{7}{c}{T1}
\\
\multirow{1}{*}{\Rmnum{1}} & $\bigcirc$ & & & 0.1740 & 0.2310 & \textbf{0.2376} \\
\cdashline{1-7}
\multirow{2}{*}{\Rmnum{2}} & & $\bigcirc$ & & 0.1819 & 0.2074 & 0.2007 \\
 & & & $\bigcirc$(O) & \textbf{0.2184} & 0.2238 & 0.2170 \\
\cdashline{1-7}
\multirow{1}{*}{\Rmnum{3}} & $\bigcirc$ & $\bigcirc$ & & 0.1767 & 0.2257 & 0.2307 \\
\cdashline{1-7}
\multirow{3}{*}{\Rmnum{4}} & $\bigcirc$ & & $\bigcirc$(O) & 0.1888 & \textbf{0.2334} & 0.2218 \\
 & & $\bigcirc$ & $\bigcirc$(O) & 0.2001 & 0.2249 & 0.2218 \\
 & $\bigcirc$ & $\bigcirc$ & $\bigcirc$(O) & 0.1871 & 0.2329 & 0.2307 \\ 
\hline
\multicolumn{7}{c}{T2}
\\
\multirow{1}{*}{\Rmnum{1}} & $\bigcirc$ & & & 0.1616 & 0.2189 & 0.2340 \\
\cdashline{1-7}
\multirow{2}{*}{\Rmnum{2}} & & $\bigcirc$ & & 0.1769 & 0.2087 & 0.1952 \\
 & & & $\bigcirc$(O) & \textbf{0.2136} & 0.2122 & 0.2095 \\
\cdashline{1-7}
\multirow{1}{*}{\Rmnum{3}} & $\bigcirc$ & $\bigcirc$ & & 0.1667 & \textbf{0.2369} & 0.2341 \\
\cdashline{1-7}
\multirow{3}{*}{\Rmnum{4}} & $\bigcirc$ & & $\bigcirc$(O) & 0.1789 & 0.2293 & 0.2194 \\
 & & $\bigcirc$ & $\bigcirc$(O) & 0.1953 & 0.2253 & 0.2252 \\
 & $\bigcirc$ & $\bigcirc$ & $\bigcirc$(O) & 0.1784 & 0.2309 & \textbf{0.2372} \\
\hline
\end{tabular}\\
\end{center}
\end{table}

\subsection{Q2: Performance of Different Aggregators}
We also compare different types of aggregators on the raw crowd answers. We use the mean results of crowd aggregators and LLM aggregators in Table \ref{tab:singleworkerresults} and add the results of the model aggregators to list them in Table \ref{tab:aggregatorperformance}. The dataset-wise model aggregator RASA that models worker reliability performs best in many cases, and performs better than instance-wise model aggregators SMV and SMS.  
LLM aggregators sometimes perform best because they not only select answers from the raw answers but also infer answers through their language understanding and reasoning ability. 

\subsection{Q3: Results on Text Answer Aggregation}
Table \ref{tab:aggresults_gleu} shows the results of text answer aggregation on the evaluation metric GLEU. $\mathcal{A}_{L.A.}$ with (O) uses GPT-4. 
To facilitate the discussions, the results are organized into four groups. Group \Rmnum{1} represents the existing work with only crowd creators and the single-stage framework; Group \Rmnum{2} is the performance of our CAMS framework with only crowd aggregators or LLM aggregators; Group \Rmnum{3} and Group \Rmnum{4} are answer aggregations with our CAMS framework with more than two answer resources; Group \Rmnum{4} includes the human-LLM hybrid combinations with LLM aggregators. \textit{Groups \Rmnum{2}, \Rmnum{3} and \Rmnum{4} are our proposals}. 
We address the major observations as follows. 

First, comparing the results among three columns, SMV performs worst in all three model aggregators. It shows that utilizing a strong model aggregator rather than a majority-based na\"ive aggregator is important for answer aggregation. Although the L.A.-only version of our approach performs best in all three datasets when using SMV, the performance is always lower than those using SMS and RASA. The following discussions will only focus on the results based on SMS and RASA. 

\begin{table}[!t]
\scriptsize
\caption{\label{tab:aggresults_meteor} Results of text answer aggregation: METEOR; LLM aggregator (GPT-4 (O)).} 
\begin{center}
\begin{tabular}{cccc|c|cc}
\hline
\multicolumn{4}{c|}{Answers} & \multicolumn{3}{c}{Model Aggregator} \\
 & $\mathcal{A}_{C.C.}$ & $\mathcal{A}_{C.A.}$ & $\mathcal{A}_{L.A.}$ & SMV & SMS & RASA \\ 
\hline
\multicolumn{7}{c}{J1}
\\
\multirow{1}{*}{\Rmnum{1}} & $\bigcirc$ & & & 0.3861 & 0.4666 & 0.4745 \\ \cdashline{1-7}
\multirow{2}{*}{\Rmnum{2}} & & $\bigcirc$ & & 0.4397 & 0.4912 & 0.4524 \\
 & & & $\bigcirc$(O) & \textbf{0.5049} & 0.5110 & 0.5070 \\
\cdashline{1-7}
\multirow{1}{*}{\Rmnum{3}} & $\bigcirc$ & $\bigcirc$ & & 0.4040 & 0.5092 & 0.4946 \\
\cdashline{1-7}
\multirow{3}{*}{\Rmnum{4}} & $\bigcirc$ & & $\bigcirc$(O) & 0.4258 & 0.5166 & 0.5076 \\
 & & $\bigcirc$ & $\bigcirc$(O) & 0.4723 & \textbf{0.5248} & 0.5064 \\
 & $\bigcirc$ & $\bigcirc$ & $\bigcirc$(O) & 0.4293 & 0.5240 & \textbf{0.5147} \\
\hline
\multicolumn{7}{c}{T1}
\\
\multirow{1}{*}{\Rmnum{1}} & $\bigcirc$ & & & 0.3786 & 0.4548 & \textbf{0.4718} \\ \cdashline{1-7}
\multirow{2}{*}{\Rmnum{2}} & & $\bigcirc$ & & 0.3927 & 0.4330 & 0.4272 \\
 & & & $\bigcirc$(O) & \textbf{0.4294} & 0.4351 & 0.4275 \\
\cdashline{1-7}
\multirow{1}{*}{\Rmnum{3}} & $\bigcirc$ & $\bigcirc$ & & 0.3833 & 0.4528 & 0.4564 \\
\cdashline{1-7}
\multirow{3}{*}{\Rmnum{4}} & $\bigcirc$ & & $\bigcirc$(O) & 0.3955 & 0.4538 & 0.4378 \\
 & & $\bigcirc$ & $\bigcirc$(O) & 0.4110 & 0.4406 & 0.4366 \\
 & $\bigcirc$ & $\bigcirc$ & $\bigcirc$(O) & 0.3948 & \textbf{0.4609} & 0.4495 \\ 
\hline
\multicolumn{7}{c}{T2}
\\
\multirow{1}{*}{\Rmnum{1}} & $\bigcirc$ & & & 0.3604 & 0.4426 & \textbf{0.4653} \\ \cdashline{1-7}
\multirow{2}{*}{\Rmnum{2}} & & $\bigcirc$ & & 0.3746 & 0.4238 & 0.4083 \\
 & & & $\bigcirc$(O) & \textbf{0.4278} & 0.4330 & 0.4287 \\
\cdashline{1-7}
\multirow{1}{*}{\Rmnum{3}} & $\bigcirc$ & $\bigcirc$ & & 0.3651 & 0.4507 & 0.4551 \\
\cdashline{1-7}
\multirow{3}{*}{\Rmnum{4}} & $\bigcirc$ & & $\bigcirc$(O) & 0.3829 & 0.4523 & 0.4402 \\
 & & $\bigcirc$ & $\bigcirc$(O) & 0.4012 & 0.4423 & 0.4482 \\
 & $\bigcirc$ & $\bigcirc$ & $\bigcirc$(O) & 0.3808 & \textbf{0.4606} & 0.4626 \\
\hline
\end{tabular}\\
\end{center}
\end{table}

\begin{table}[t!]
\scriptsize
\caption{\label{tab:aggresults_embedding} Results of text answer aggregation: Embedding Similarity; LLM aggregator (GPT-4 (O)).}
\begin{center}
\begin{tabular}{cccc|c|cc}
\hline
\multicolumn{4}{c|}{Answers} & \multicolumn{3}{c}{Model Aggregator} \\
 & $\mathcal{A}_{C.C.}$ & $\mathcal{A}_{C.A.}$ & $\mathcal{A}_{L.A.}$ & SMV & SMS & RASA \\ 
\hline
\multicolumn{7}{c}{J1}
\\
\multirow{1}{*}{\Rmnum{1}} & $\bigcirc$ & & & 0.6732 & 0.7426 & 0.7414 \\
\cdashline{1-7}
\multirow{2}{*}{\Rmnum{2}} & & $\bigcirc$ & & 0.6983 & 0.7360 & 0.7188 \\
 & & & $\bigcirc$(O) & \textbf{0.7313} & 0.7332 & 0.7317 \\
\cdashline{1-7}
\multirow{1}{*}{\Rmnum{3}} & $\bigcirc$ & $\bigcirc$ & & 0.6815 & \textbf{0.7516} & \textbf{0.7459} \\
\cdashline{1-7}
\multirow{3}{*}{\Rmnum{4}} & $\bigcirc$ & & $\bigcirc$(O) & 0.6926 & 0.7427 & 0.7356 \\
 & & $\bigcirc$ & $\bigcirc$(O) & 0.7148 & 0.7413 & 0.7336 \\
 & $\bigcirc$ & $\bigcirc$ & $\bigcirc$(O) & 0.6940 & 0.7484 & 0.7410 \\
\hline
\multicolumn{7}{c}{T1}
\\
\multirow{1}{*}{\Rmnum{1}} & $\bigcirc$ & & & 0.7245 & 0.7935 & \textbf{0.8020} \\
\cdashline{1-7}
\multirow{2}{*}{\Rmnum{2}} & & $\bigcirc$ & & 0.7430 & 0.7851 & 0.7758 \\
 & & & $\bigcirc$(O) & \textbf{0.7786} & 0.7807 & 0.7762 \\
\cdashline{1-7}
\multirow{1}{*}{\Rmnum{3}} & $\bigcirc$ & $\bigcirc$ & & 0.7306 & 0.7965 & 0.7976 \\
\cdashline{1-7}
\multirow{3}{*}{\Rmnum{4}} & $\bigcirc$ & & $\bigcirc$(O) & 0.7425 & 0.7894 & 0.7793 \\
 & & $\bigcirc$ & $\bigcirc$(O) & 0.7608 & 0.7854 & 0.7807 \\
 & $\bigcirc$ & $\bigcirc$ & $\bigcirc$(O) & 0.7426 & \textbf{0.7993} & 0.7890 \\
\hline
\multicolumn{7}{c}{T2}
\\
\multirow{1}{*}{\Rmnum{1}} & $\bigcirc$ & & & 0.7105 & 0.7860 & \textbf{0.7935} \\ \cdashline{1-7}
\multirow{2}{*}{\Rmnum{2}} &  & $\bigcirc$ & & 0.7282 & 0.7681 & 0.7468 \\
 & & & $\bigcirc$(O) & \textbf{0.7651} & 0.7725 & 0.7673 \\
\cdashline{1-7}
\multirow{1}{*}{\Rmnum{3}} & $\bigcirc$ & $\bigcirc$ & & 0.7164 & \textbf{0.7885} & 0.7822 \\
\cdashline{1-7}
\multirow{3}{*}{\Rmnum{4}} & $\bigcirc$ & & $\bigcirc$(O) & 0.7287 & 0.7856 & 0.7744 \\
 & & $\bigcirc$ & $\bigcirc$(O) & 0.7467 & 0.7712 & 0.7728 \\
 & $\bigcirc$ & $\bigcirc$ & $\bigcirc$(O) & 0.7286 & 0.7851 & 0.7824 \\
\hline
\end{tabular}\\
\end{center}
\end{table}

Second, comparing the SMS (RASA) results between the L.A.-only version in Group \Rmnum{2} and Group \Rmnum{4}, the versions in Group \Rmnum{4} that utilize human-LLM hybrid aggregation outperform the version in Group \Rmnum{2} that only utilizes the answers of the LLM aggregators in most of the cases. In addition, the versions in Group \Rmnum{4} also always outperform the results of LLM aggregators without model aggregators. They show that, although LLM aggregators perform better than the average of crowd workers in Table \ref{tab:singleworkerresults}, the answers provided by crowd workers contain valuable information for model aggregators to generate good answers. One purpose of answer aggregation methods is to estimate good workers and good answers from the raw crowd answers (looking for the needle in a haystack). Following the observation in Table \ref{tab:singleworkerresults} that there are still good crowd workers (creators and aggregators) who perform better than LLMs (aggregators), a strong model aggregator can target such good workers and answers. In contrast to replacing crowd workers by only using LLMs, human-LLM hybrid solutions potentially have advantages. 

Third, Table \ref{tab:aggresults_meteor} and \ref{tab:aggresults_embedding} list the results on the evaluation metrics METEOR and Embedding Similarity. They reach similar observations with the evaluation metric GLEU.  
 
Fourth, regarding the performance of different components (ablations) of the proposed method within the CAMS framework, (1) the results of columns C.A. or L.A. in Table \ref{tab:aggregatorperformance} represent components that include only crowd aggregators or LLM aggregators, without the use of model aggregators; (2) Group \Rmnum{1} represents components with only crowd creators and the single-stage framework. Comparing these results with those of Groups \Rmnum{2}, \Rmnum{3} and Group \Rmnum{4} in Tables \ref{tab:aggresults_gleu}, \ref{tab:aggresults_meteor} and \ref{tab:aggresults_embedding} illustrates the effectiveness of our human-LLM hybrid aggregation approach and CAMS framework. 

\begin{figure*}[!t]
\begin{minipage}{0.315\textwidth}
\centering
\includegraphics[height=3.5cm]{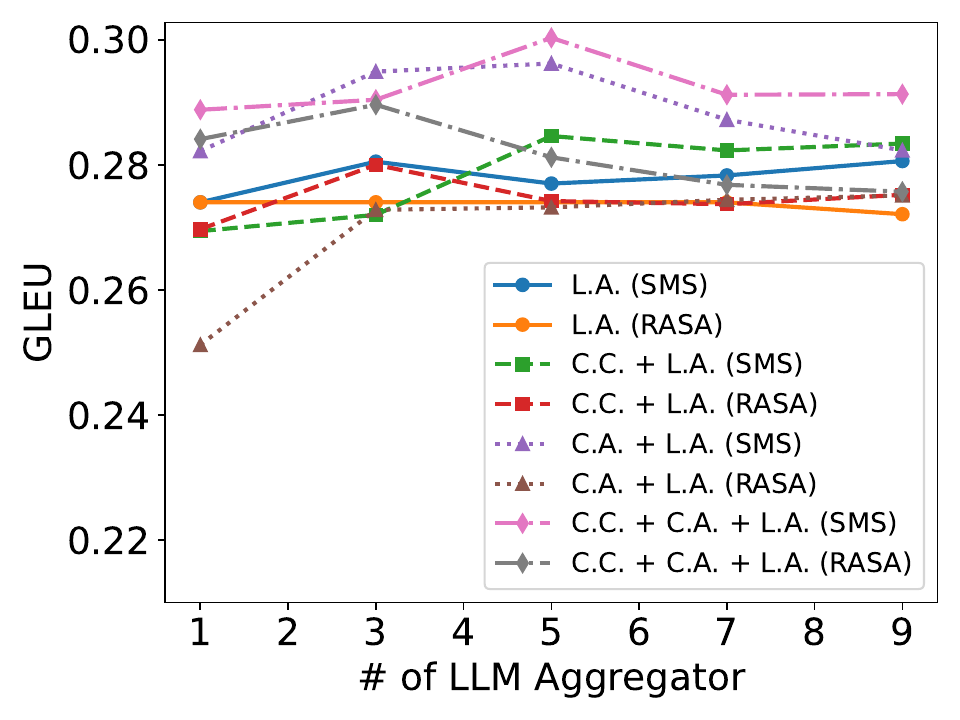}
\\(a) J1
\end{minipage}
\begin{minipage}{0.315\textwidth}
\centering
\includegraphics[height=3.5cm]{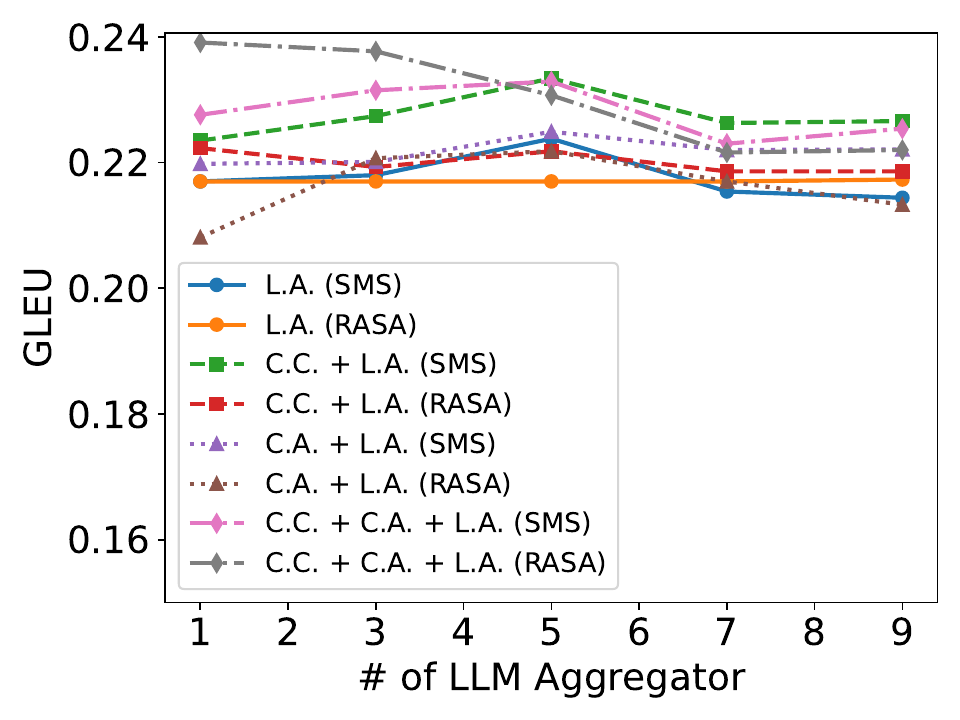}
\\(b) T1
\end{minipage}
\begin{minipage}{0.315\textwidth}
\centering
\includegraphics[height=3.5cm]{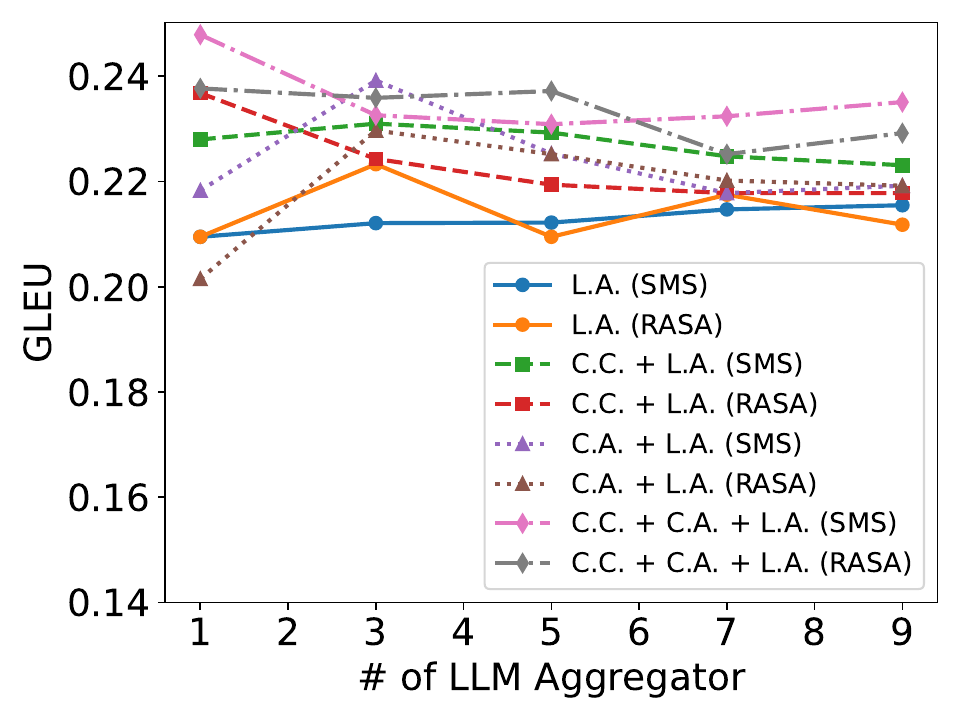}
\\(c) T2
\end{minipage}
\caption{GLEU Results by different number of L.A. (GPT-4 (O)) based on SMS and RASA. }
\label{fig:diff_num_llmagg_gleu_chatgpt} 
\end{figure*}

\begin{figure*}[!t]
\begin{minipage}{0.315\textwidth}
\centering
\includegraphics[height=3.5cm]{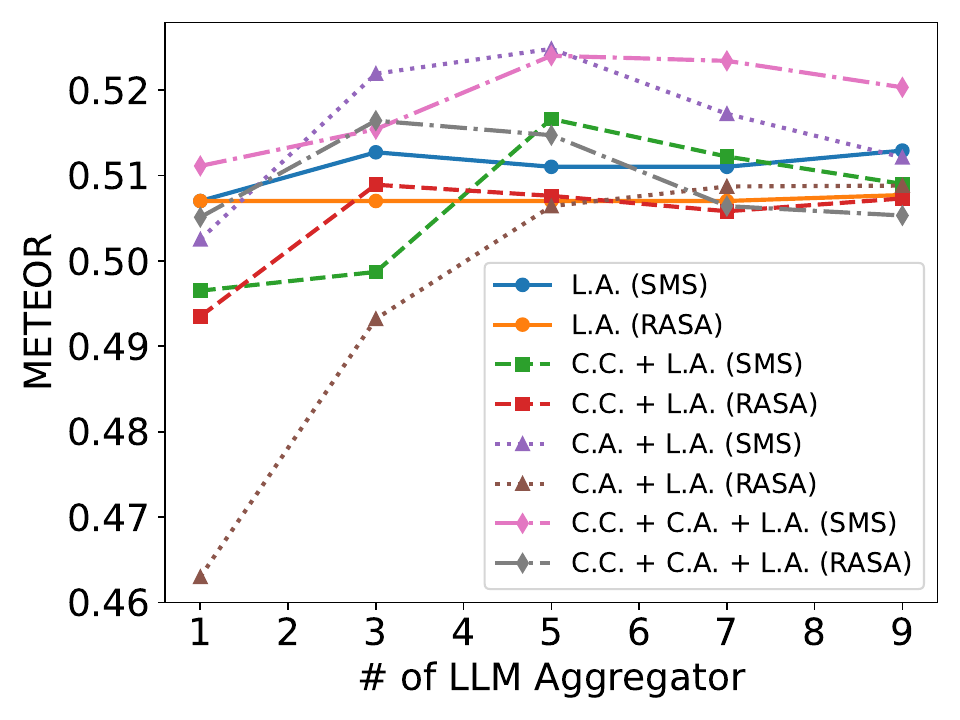}
\\(a) J1
\end{minipage}
\begin{minipage}{0.315\textwidth}
\centering
\includegraphics[height=3.5cm]{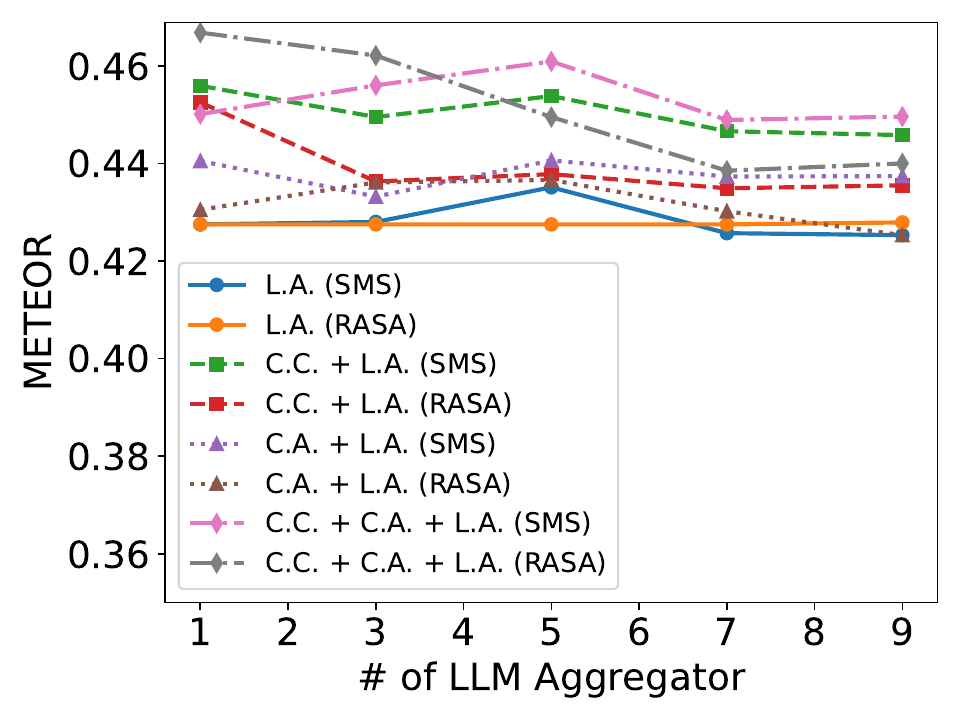}
\\(b) T1
\end{minipage}
\begin{minipage}{0.315\textwidth}
\centering
\includegraphics[height=3.5cm]{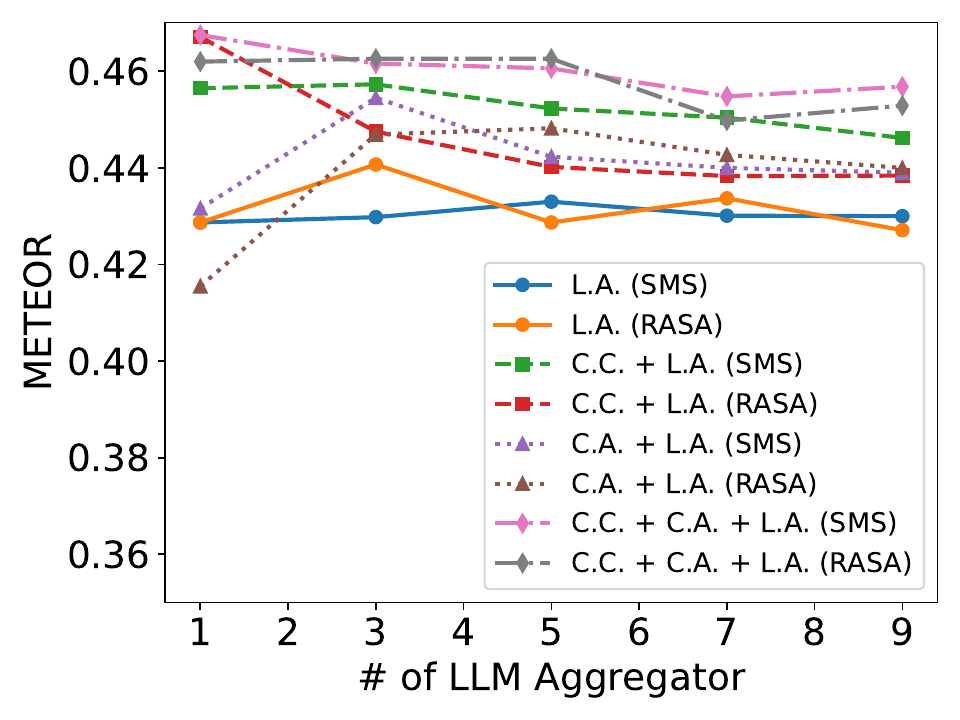}
\\(c) T2
\end{minipage}
\caption{METEOR Results by different number of L.A. (GPT-4 (O)) based on SMS and RASA. }
\label{fig:diff_num_llmagg_meteor_chatgpt} 
\end{figure*}

\begin{figure*}[!t]
\begin{minipage}{0.315\textwidth}
\centering
\includegraphics[height=3.5cm]{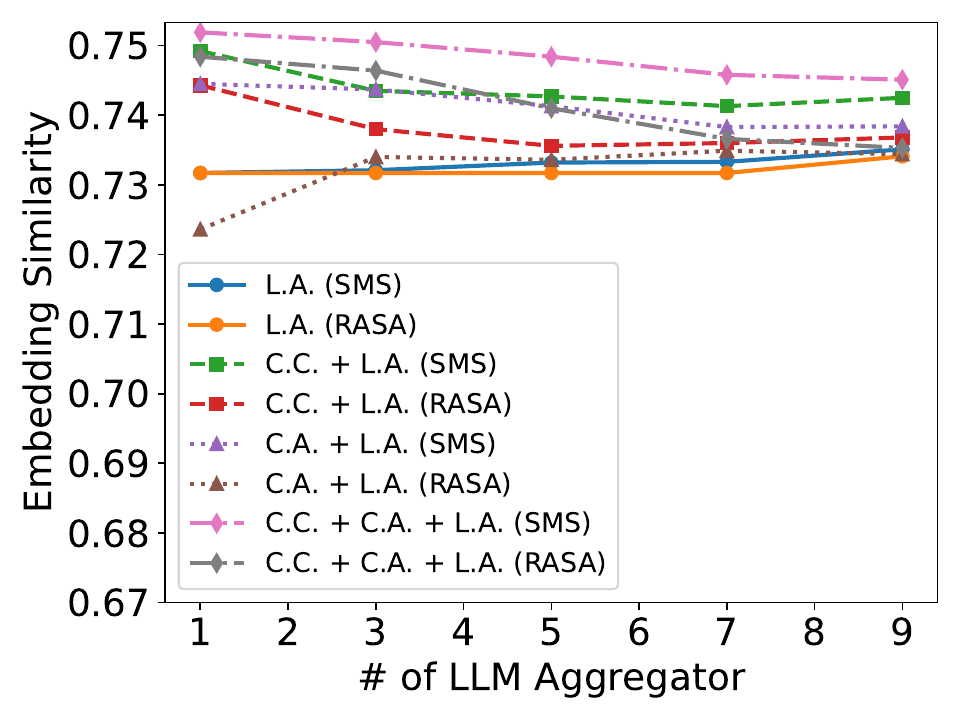}
\\(a) J1
\end{minipage}
\begin{minipage}{0.315\textwidth}
\centering
\includegraphics[height=3.5cm]{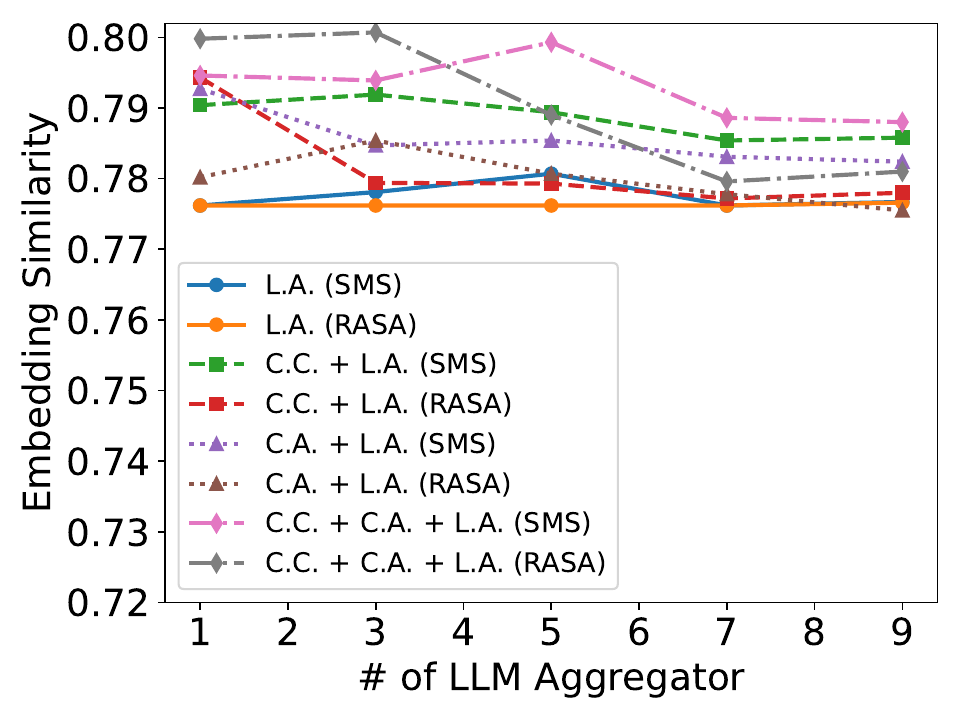}
\\(b) T1
\end{minipage}
\begin{minipage}{0.315\textwidth}
\centering
\includegraphics[height=3.5cm]{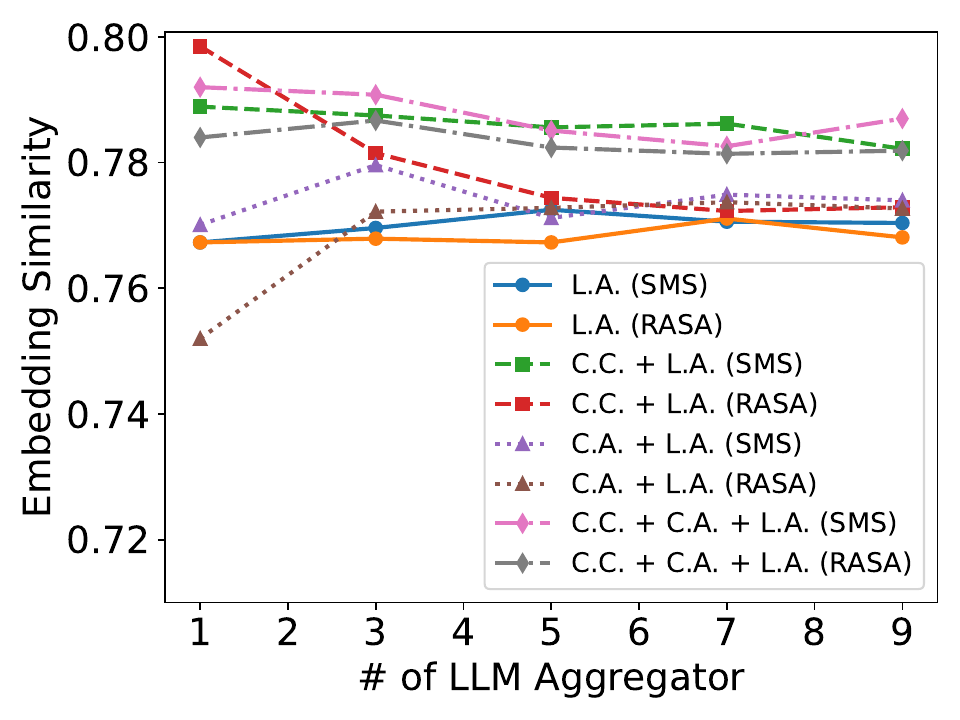}
\\(c) T2
\end{minipage}
\caption{Embedding Similarity Results by different number of L.A. (GPT-4 (O)) based on SMS and RASA. }
\label{fig:diff_num_llmagg_embedsim_chatgpt} 
\end{figure*}

Fifth, in the main results, we utilize the same number of LLM aggregators with crowd aggregators, we thus also investigate the influences of different numbers of LLM aggregators on the performance. Figures \ref{fig:diff_num_llmagg_gleu_chatgpt}, \ref{fig:diff_num_llmagg_meteor_chatgpt} and \ref{fig:diff_num_llmagg_embedsim_chatgpt} show the results on the evaluation metrics GLEU, METEOR and Embedding Similarity by GPT-4 LLM aggregators based on model aggregators SMS and RASA. The results by Gemini LLM aggregators are in the Appendix. No consistent patterns were observed across different datasets; sometimes, more LLM aggregators are better; sometimes, fewer LLM aggregators are better; it depends on the datasets; we cannot know which case it is in practical crowd annotations. In all cases of the J1 dataset and some cases of the T1 and T2 datasets, the results of only using one LLM aggregator are bad; using multiple LLM aggregators is recommended to avoid risk.  

Finally, the approach we propose in this work has several versions (six combinations of worker and answer resources in Groups \Rmnum{2}, \Rmnum{3}, and \Rmnum{4}). The best version depends on the specific application. If there is no prior knowledge of the target task or dataset, we recommend using all workers and answers (i.e., crowd creators, crowd aggregators, and LLM aggregators) together with a reliable model aggregator (e.g., SMS or RASA).

\section{Conclusion}
In this paper, we investigate the capability of LLMs as aggregators in the scenario of crowd text answer aggregation for the quality control of crowdsourced close-ended text answer annotation tasks. We propose a human-LLM hybrid text answer aggregation method with a Creator-Aggregator Multi-Stage (CAMS) crowdsourcing framework. 
The experimental results show both crowd aggregators and LLM aggregators can generate estimated answers with higher quality than raw crowd answers provided by crowd creators. Our CAMS approach with model aggregators can further improve the answer quality based on the combinations of three resources of workers and answers. 
This work presents a solution based on human-AI collaboration for a specific task. Humans can leverage AI to enhance the quality and efficiency of tasks like data annotation. 
We share the datasets and source codes at \url{https://github.com/garfieldpigljy/HumanLLMHybridAggregation}.

\section*{Limitations}
In this work, we only verify two popular commercial LLMs, i.e., GPT-4 and Gemini-1.5-Pro. 
We think that GPT-4 and Gemini-1.5-Pro can represent the performance that LLMs can currently achieve, and they are enough for the empirical experiments. On the other hand, the performance of other LLMs (e.g., some open-source LLM models) is also an interesting issue for investigation. 
In the future work, we will verify other LLMs.  

For the purpose of investigating the results with one LLM, people usually select a single LLM provider to complete the tasks. In the experiments, we used different LLMs (GPT-4 or Gemini) separately. Different LLM aggregators are constructed by using different temperature parameters. As shown in the experimental results in Section \ref{sec:experiments_individual}, the answers generated by multiple LLM aggregators constructed by the same LLM have relatively low diversity. In future work, a candidate solution is using different LLMs and different temperatures to construct many LLM aggregators. 

\section*{Acknowledgements}
This work was supported by JSPS KAKENHI Grant Number JP23K28092 and JP23K11227. 

\section*{Ethic Statement}
No personal information is used in this study, and therefore no demographic or geographic characteristics of the annotators are included or utilized in the data. The crowd workers have obtained sufficient wages by local standards.

\bibliography{contents/llm,contents/crowd}

\begin{thebibliography}{55}
\expandafter\ifx\csname natexlab\endcsname\relax\def\natexlab#1{#1}\fi

\bibitem[{Baba and Kashima(2013)}]{twostagequality}
Yukino Baba and Hisashi Kashima. 2013.
\newblock \href {https://doi.org/10.1145/2487575.2487600} {Statistical quality estimation for general crowdsourcing tasks}.
\newblock In \emph{Proceedings of the 19th ACM SIGKDD International Conference on Knowledge Discovery and Data Mining (KDD)}, pages 554--562.

\bibitem[{Bachrach et~al.(2012)Bachrach, Minka, Guiver, and Graepel}]{DARE}
Yoram Bachrach, Tom Minka, John Guiver, and Thore Graepel. 2012.
\newblock \href {http://dl.acm.org/citation.cfm?id=3042573.3042680} {How to grade a test without knowing the answers: A bayesian graphical model for adaptive crowdsourcing and aptitude testing}.
\newblock In \emph{Proceedings of the 29th International Coference on International Conference on Machine Learning (ICML)}, pages 819--826.

\bibitem[{Bernstein et~al.(2015)Bernstein, Little, Miller, Hartmann, Ackerman, Karger, Crowell, and Panovich}]{findfixverify}
Michael~S. Bernstein, Greg Little, Robert~C. Miller, Bj\"{o}rn Hartmann, Mark~S. Ackerman, David~R. Karger, David Crowell, and Katrina Panovich. 2015.
\newblock Soylent: a word processor with a crowd inside.
\newblock \emph{Commun. ACM}, 58(8):85--94.

\bibitem[{Cattelan(2012)}]{preferencesurvey}
Manuela Cattelan. 2012.
\newblock Models for paired comparison data: A review with emphasis on dependent data.
\newblock \emph{Statistical Science}, pages 412--433.

\bibitem[{Causeur and Husson(2005)}]{2dbt}
David Causeur and Fran{\c{c}}ois Husson. 2005.
\newblock A 2-dimensional extension of the bradley--terry model for paired comparisons.
\newblock \emph{Journal of Statistical Planning and Inference}, 135(2):245--259.

\bibitem[{Cegin et~al.(2023)Cegin, Simko, and Brusilovsky}]{CrowdChatgptIntentClassification}
Jan Cegin, Jakub Simko, and Peter Brusilovsky. 2023.
\newblock \href {https://doi.org/10.18653/v1/2023.emnlp-main.117} {{C}hat{GPT} to replace crowdsourcing of paraphrases for intent classification: Higher diversity and comparable model robustness}.
\newblock In \emph{Proceedings of the 2023 Conference on Empirical Methods in Natural Language Processing (EMNLP)}, pages 1889--1905.

\bibitem[{Cer et~al.(2018)Cer, Yang, Kong, and et~al.}]{USE}
Daniel Cer, Yinfei Yang, Sheng{-}yi Kong, and et~al. 2018.
\newblock \href {http://arxiv.org/abs/1803.11175} {Universal sentence encoder}.
\newblock \emph{CoRR}, abs/1803.11175.

\bibitem[{Chen and Joachims(2016{\natexlab{a}})}]{ChenIntransitivityWSDMlong}
Shuo Chen and Thorsten Joachims. 2016{\natexlab{a}}.
\newblock \href {https://doi.org/10.1145/2835776.2835787} {Modeling intransitivity in matchup and comparison data}.
\newblock In \emph{Proceedings of the Ninth ACM International Conference on Web Search and Data Mining (WSDM)}, pages 227--236.

\bibitem[{Chen and Joachims(2016{\natexlab{b}})}]{ChenIntransitivityKDDlong}
Shuo Chen and Thorsten Joachims. 2016{\natexlab{b}}.
\newblock \href {https://doi.org/10.1145/2939672.2939764} {Predicting matchups and preferences in context}.
\newblock In \emph{Proceedings of the 22nd ACM SIGKDD International Conference on Knowledge Discovery and Data Mining (KDD)}, pages 775--784.

\bibitem[{Chen et~al.(2013)Chen, Bennett, Collins-Thompson, and Horvitz}]{crowdbtlong}
Xi~Chen, Paul~N. Bennett, Kevyn Collins-Thompson, and Eric Horvitz. 2013.
\newblock \href {https://doi.org/10.1145/2433396.2433420} {Pairwise ranking aggregation in a crowdsourced setting}.
\newblock In \emph{Proceedings of the 6th ACM International Conference on Web Search and Data Mining (WSDM)}, pages 193--202.

\bibitem[{Cheng et~al.(2015)Cheng, Teevan, Iqbal, and Bernstein}]{PairCompareSort}
Justin Cheng, Jaime Teevan, Shamsi~T. Iqbal, and Michael~S. Bernstein. 2015.
\newblock \href {https://doi.org/10.1145/2702123.2702146} {Break it down: A comparison of macro- and microtasks}.
\newblock In \emph{Proceedings of the 33rd Annual ACM Conference on Human Factors in Computing Systems (CHI)}, pages 4061--4064.

\bibitem[{Gilardi et~al.(2023)Gilardi, Alizadeh, and Kubli}]{CrowdChatgptTextAnnotation}
Fabrizio Gilardi, Meysam Alizadeh, and Maël Kubli. 2023.
\newblock \href {https://doi.org/10.1073/pnas.2305016120} {Chatgpt outperforms crowd workers for text-annotation tasks}.
\newblock \emph{Proceedings of the National Academy of Sciences}, 120(30):e2305016120.

\bibitem[{Gomes et~al.(2011)Gomes, Welinder, Krause, and Perona}]{crowdclustering}
Ryan~G. Gomes, Peter Welinder, Andreas Krause, and Pietro Perona. 2011.
\newblock Crowdclustering.
\newblock In \emph{Advances in Neural Information Processing Systems 24 (NIPS)}, pages 558--566.

\bibitem[{Hartmann and Sonntag(2022)}]{ExplainNLPSurvey}
Mareike Hartmann and Daniel Sonntag. 2022.
\newblock A survey on improving {NLP} models with human explanations.
\newblock In \emph{Proceedings of the First Workshop on Learning with Natural Language Supervision}, pages 40--47.

\bibitem[{He et~al.(2024{\natexlab{a}})He, Lin, Gong, and et~al.}]{annollm}
Xingwei He, Zhenghao Lin, Yeyun Gong, and et~al. 2024{\natexlab{a}}.
\newblock \href {https://doi.org/10.18653/v1/2024.naacl-industry.15} {{A}nno{LLM}: Making large language models to be better crowdsourced annotators}.
\newblock In \emph{Proceedings of the 2024 Conference of the North American Chapter of the Association for Computational Linguistics: Human Language Technologies (Volume 6: Industry Track) (NAACL)}, pages 165--190.

\bibitem[{He et~al.(2024{\natexlab{b}})He, Huang, Ding, Rohatgi, and Huang}]{llmcrowdaggCHI24}
Zeyu He, Chieh-Yang Huang, Chien-Kuang~Cornelia Ding, Shaurya Rohatgi, and Ting-Hao~Kenneth Huang. 2024{\natexlab{b}}.
\newblock \href {https://doi.org/10.1145/3613904.3642834} {If in a crowdsourced data annotation pipeline, a gpt-4}.
\newblock In \emph{Proceedings of the CHI Conference on Human Factors in Computing Systems (CHI)}.

\bibitem[{Hinton and Roweis(2002)}]{sne}
Geoffrey~E Hinton and Sam~T. Roweis. 2002.
\newblock Stochastic neighbor embedding.
\newblock In \emph{Advances in Neural Information Processing Systems 15 (NIPS)}, pages 857--864.

\bibitem[{Jin et~al.(2020)Jin, Xu, Gu, and Farnoud}]{HBTL}
Tao Jin, Pan Xu, Quanquan Gu, and Farzad Farnoud. 2020.
\newblock \href {https://doi.org/10.1609/aaai.v34i04.5860} {Rank aggregation via heterogeneous thurstone preference models}.
\newblock In \emph{Proceedings of the AAAI Conference on Artificial Intelligence (AAAI)}, volume~34, pages 4353--4360.

\bibitem[{Kawase et~al.(2019)Kawase, Kuroki, and Miyauchi}]{graphexpert}
Yasushi Kawase, Yuko Kuroki, and Atsushi Miyauchi. 2019.
\newblock \href {https://doi.org/10.24963/ijcai.2019/177} {Graph mining meets crowdsourcing: Extracting experts for answer aggregation}.
\newblock In \emph{Proceedings of the Twenty-Eighth International Joint Conference on Artificial Intelligence (IJCAI)}, pages 1272--1279.

\bibitem[{Kittur et~al.(2011)Kittur, Smus, Khamkar, and Kraut}]{PartitionMapReduce}
Aniket Kittur, Boris Smus, Susheel Khamkar, and Robert~E. Kraut. 2011.
\newblock \href {https://doi.org/10.1145/2047196.2047202} {Crowdforge: crowdsourcing complex work}.
\newblock In \emph{Proceedings of the 24th Annual ACM Symposium on User Interface Software and Technology (UIST)}, pages 43--52.

\bibitem[{Kobayashi(2018)}]{ModelEnsembleSummary}
Hayato Kobayashi. 2018.
\newblock \href {https://www.aclweb.org/anthology/D18-1449} {Frustratingly easy model ensemble for abstractive summarization}.
\newblock In \emph{Proceedings of the 2018 Conference on Empirical Methods in Natural Language Processing (EMNLP)}, pages 4165--4176.

\bibitem[{Kulkarni et~al.(2012)Kulkarni, Can, and Hartmann}]{pricedividesolve}
Anand Kulkarni, Matthew Can, and Bj\"{o}rn Hartmann. 2012.
\newblock \href {https://doi.org/10.1145/2145204.2145354} {Collaboratively crowdsourcing workflows with turkomatic}.
\newblock In \emph{Proceedings of the ACM 2012 Conference on Computer Supported Cooperative Work (CSCW)}, pages 1003--1012.

\bibitem[{Li et~al.(2022)Li, Sun, and Li}]{lfcxfeature}
Jingzheng Li, Hailong Sun, and Jiyi Li. 2022.
\newblock \href {https://doi.org/10.1007/s10994-022-06211-x} {Beyond confusion matrix: Learning from multiple annotators with awareness of instance features}.
\newblock \emph{Mach. Learn.}, 112(3):1053--1075.

\bibitem[{Li(2019)}]{budgetcost}
Jiyi Li. 2019.
\newblock \href {https://doi.org/10.1007/978-3-030-36802-9\_26} {Budget cost reduction for label collection with confusability based exploration}.
\newblock In \emph{Proceedings of the 26th International Conference on Neural Information Processing (ICONIP)}, pages 231--241.

\bibitem[{Li(2020)}]{textaggregation}
Jiyi Li. 2020.
\newblock \href {https://doi.org/10.1145/3397271.3401239} {Crowdsourced text sequence aggregation based on hybrid reliability and representation}.
\newblock In \emph{Proceedings of the 43rd International ACM SIGIR Conference on Research and Development in Information Retrieval (SIGIR)}, pages 1761--1764.

\bibitem[{Li(2022)}]{contextpairwisecrowdopinions}
Jiyi Li. 2022.
\newblock \href {https://doi.org/10.1145/3485447.3512137} {Context-based collective preference aggregation for prioritizing crowd opinions in social decision-making}.
\newblock In \emph{Proceedings of the ACM Web Conference 2022 (WWW)}, pages 2657--2667.

\bibitem[{Li(2024)}]{llmcrowdaggICASSP2024}
Jiyi Li. 2024.
\newblock \href {https://doi.org/10.1109/ICASSP48485.2024.10447803} {A comparative study on annotation quality of crowdsourcing and llm via label aggregation}.
\newblock In \emph{2024 IEEE International Conference on Acoustics, Speech and Signal Processing (ICASSP)}, pages 6525--6529.

\bibitem[{Li et~al.(2018{\natexlab{a}})Li, Baba, and Kashima}]{workersimilarity}
Jiyi Li, Yukino Baba, and Hisashi Kashima. 2018{\natexlab{a}}.
\newblock \href {https://doi.org/10.1007/978-3-030-01421-6\_57} {Incorporating worker similarity for label aggregation in crowdsourcing}.
\newblock In \emph{Proceedings of the 27th International Conference on Artificial Neural Networks (ICANN)}, pages 596--606.

\bibitem[{Li et~al.(2018{\natexlab{b}})Li, Baba, and Kashima}]{rankingclusteringlong}
Jiyi Li, Yukino Baba, and Hisashi Kashima. 2018{\natexlab{b}}.
\newblock \href {https://doi.org/10.24963/ijcai.2018/215} {Simultaneous clustering and ranking from pairwise comparisons}.
\newblock In \emph{Proceedings of the Twenty-Seventh International Joint Conference on Artificial Intelligence (IJCAI)}, pages 1554--1560.

\bibitem[{Li et~al.(2021)Li, Endo, and Kashima}]{crowdtriplet}
Jiyi Li, Lucas~Ryo Endo, and Hisashi Kashima. 2021.
\newblock Label aggregation for crowdsourced triplet similarity comparisons.
\newblock In \emph{Proceedings of the 28th International Conference on Neural Information Processing (ICONIP)}, pages 176--185.

\bibitem[{Li and Fukumoto(2019)}]{textaggregationdataset}
Jiyi Li and Fumiyo Fukumoto. 2019.
\newblock \href {https://doi.org/10.18653/v1/D19-5904} {A dataset of crowdsourced word sequences: Collections and answer aggregation for ground truth creation}.
\newblock In \emph{Proceedings of the First Workshop on Aggregating and Analysing Crowdsourced Annotations for NLP (AnnoNLP2019)}, pages 24--28.

\bibitem[{Li and Kashima(2017)}]{IterativeReduction}
Jiyi Li and Hisashi Kashima. 2017.
\newblock \href {https://doi.org/10.1145/978-3-319-68786-5\_4} {Iterative reduction worker filtering for crowdsourced label aggregation}.
\newblock In \emph{Proceedings of the 18th International Conference on Web Information Systems Engineering (WISE)}, pages 46--54.

\bibitem[{Li et~al.(2020)Li, Kawase, Baba, and Kashima}]{crowdpr}
Jiyi Li, Yasushi Kawase, Yukino Baba, and Hisashi Kashima. 2020.
\newblock Performance as a constraint: An improved wisdom of crowds using performance regularization.
\newblock In \emph{Proceedings of the Twenty-Ninth International Joint Conference on Artificial Intelligence (IJCAI)}, pages 1534--1541.

\bibitem[{Li et~al.(2014)Li, Li, Gao, Su, Zhao, Demirbas, Fan, and Han}]{catd}
Qi~Li, Yaliang Li, Jing Gao, Lu~Su, Bo~Zhao, Murat Demirbas, Wei Fan, and Jiawei Han. 2014.
\newblock \href {https://doi.org/10.14778/2735496.2735505} {A confidence-aware approach for truth discovery on long-tail data}.
\newblock \emph{Proc. VLDB Endow.}, 8(4):425--436.

\bibitem[{Lu et~al.(2023)Lu, Li, Takeuchi, and Kashima}]{multiviewcrowdtriplet}
Xiaotian Lu, Jiyi Li, Koh Takeuchi, and Hisashi Kashima. 2023.
\newblock \href {https://doi.org/10.1145/3543507.3583431} {Multiview representation learning from crowdsourced triplet comparisons}.
\newblock In \emph{Proceedings of the ACM Web Conference 2023 (WWW)}, pages 3827--3836.

\bibitem[{Oh et~al.(2015)Oh, Thekumparampil, and Xu}]{matrix2long}
Sewoong Oh, Kiran~K Thekumparampil, and Jiaming Xu. 2015.
\newblock Collaboratively learning preferences from ordinal data.
\newblock In \emph{Advances in Neural Information Processing Systems 28 (NIPS)}, pages 1909--1917.

\bibitem[{Raman and Joachims(2014)}]{peergrading}
Karthik Raman and Thorsten Joachims. 2014.
\newblock \href {https://doi.org/10.1145/2623330.2623654} {Methods for ordinal peer grading}.
\newblock In \emph{Proceedings of the 20th ACM SIGKDD International Conference on Knowledge Discovery and Data Mining (KDD)}, pages 1037--1046.

\bibitem[{Snow et~al.(2008)Snow, O'Connor, Jurafsky, and Ng}]{mj}
R.~Snow, B.~O'Connor, D.~Jurafsky, and A.~Y. Ng. 2008.
\newblock Cheap and fast---but is it good?: Evaluating non-expert annotations for natural language tasks.
\newblock In \emph{Proceedings of the Conference on Empirical Methods in Natural Language Processing (EMNLP)}, pages 254--263.

\bibitem[{T{\"{o}}rnberg(2023)}]{CrowdChatgptTwitter}
Petter T{\"{o}}rnberg. 2023.
\newblock \href {https://doi.org/10.48550/ARXIV.2304.06588} {Chatgpt-4 outperforms experts and crowd workers in annotating political twitter messages with zero-shot learning}.
\newblock \emph{CoRR}, abs/2304.06588.

\bibitem[{van~der Maaten and Hinton(2008)}]{tsnelong}
Laurens van~der Maaten and Geoffrey Hinton. 2008.
\newblock Visualizing data using t-{SNE}.
\newblock \emph{Journal of Machine Learning Research}, 9(Nov):2579--2605.

\bibitem[{van~der Maaten and Weinberger(2012)}]{stelong}
Laurens van~der Maaten and Kilian Weinberger. 2012.
\newblock \href {https://doi.org/10.1109/MLSP.2012.6349720} {Stochastic triplet embedding}.
\newblock In \emph{2012 IEEE International Workshop on Machine Learning for Signal Processing (MLSP)}, pages 1--6.

\bibitem[{Venanzi et~al.(2014)Venanzi, Guiver, Kazai, Kohli, and Shokouhi}]{cbcclong}
Matteo Venanzi, John Guiver, Gabriella Kazai, Pushmeet Kohli, and Milad Shokouhi. 2014.
\newblock \href {https://doi.org/10.1145/2566486.2567989} {Community-based bayesian aggregation models for crowdsourcing}.
\newblock In \emph{Proceedings of the 23rd International Conference on World Wide Web (WWW)}, pages 155--164.

\bibitem[{Veselovsky et~al.(2023)Veselovsky, Ribeiro, and West}]{veselovsky2023artificial}
Veniamin Veselovsky, Manoel~Horta Ribeiro, and Robert West. 2023.
\newblock \href {https://doi.org/10.48550/ARXIV.2306.07899} {Artificial artificial artificial intelligence: Crowd workers widely use large language models for text production tasks}.
\newblock \emph{CoRR}, abs/2306.07899.

\bibitem[{Wauthier and Jordan(2011)}]{wauthier2011bayesian}
Fabian~L Wauthier and Michael Jordan. 2011.
\newblock \href {https://proceedings.neurips.cc/paper_files/paper/2011/file/0768281a05da9f27df178b5c39a51263-Paper.pdf} {Bayesian bias mitigation for crowdsourcing}.
\newblock In \emph{Advances in Neural Information Processing Systems (NIPS)}, volume~24.

\bibitem[{Whitehill et~al.(2009)Whitehill, Ruvolo, Wu, Bergsma, and Movellan}]{gladlong}
Jacob Whitehill, Paul Ruvolo, Tingfan Wu, Jacob Bergsma, and Javier Movellan. 2009.
\newblock Whose vote should count more: Optimal integration of labels from labelers of unknown expertise.
\newblock In \emph{Proceedings of the 22nd International Conference on Neural Information Processing Systems (NIPS)}, pages 2035--2043.

\bibitem[{Wiegreffe and Marasovic(2021)}]{ExplainNLPData}
Sarah Wiegreffe and Ana Marasovic. 2021.
\newblock \href {https://datasets-benchmarks-proceedings.neurips.cc/paper_files/paper/2021/file/698d51a19d8a121ce581499d7b701668-Paper-round1.pdf} {Teach me to explain: A review of datasets for explainable natural language processing}.
\newblock In \emph{Proceedings of the Neural Information Processing Systems Track on Datasets and Benchmarks}, volume~1.

\bibitem[{Wu et~al.(2022)Wu, Terry, and Cai}]{aichainVScrowdpipeline}
Tongshuang Wu, Michael Terry, and Carrie~Jun Cai. 2022.
\newblock \href {https://doi.org/10.1145/3491102.3517582} {Ai chains: Transparent and controllable human-ai interaction by chaining large language model prompts}.
\newblock In \emph{Proceedings of the 2022 CHI Conference on Human Factors in Computing Systems (CHI)}.

\bibitem[{Wu et~al.(2023)Wu, Zhu, Albayrak, and et~al.}]{llmcrowdpipeline}
Tongshuang Wu, Haiyi Zhu, Maya Albayrak, and et~al. 2023.
\newblock \href {https://doi.org/10.48550/ARXIV.2307.10168} {Llms as workers in human-computational algorithms? replicating crowdsourcing pipelines with llms}.
\newblock \emph{CoRR}, abs/2307.10168.

\bibitem[{Yi et~al.(2013)Yi, Jin, Jain, and Jain}]{matrix1long}
Jinfeng Yi, Rong Jin, Shaili Jain, and Anil Jain. 2013.
\newblock Inferring users’ preferences from crowdsourced pairwise comparisons: A matrix completion approach.
\newblock In \emph{Proceeding of the 1st AAAI Conference on Human Computation and Crowdsourcing (HCOMP)}.

\bibitem[{Yi et~al.(2012)Yi, Jin, Jain, Yang, and Jain}]{semicrowdclustering}
Jinfeng Yi, Rong Jin, Shaili Jain, Tianbao Yang, and Anil~K. Jain. 2012.
\newblock Semi-crowdsourced clustering: Generalizing crowd labeling by robust distance metric learning.
\newblock In \emph{Advances in Neural Information Processing Systems 25 (NIPS)}, pages 1772--1780.

\bibitem[{Yin et~al.(2017)Yin, Han, Zhang, and Yu}]{LAA}
Li'ang Yin, Jianhua Han, Weinan Zhang, and Yong Yu. 2017.
\newblock \href {https://doi.org/10.24963/ijcai.2017/184} {Aggregating crowd wisdoms with label-aware autoencoders}.
\newblock In \emph{Proceedings of the Twenty-Sixth International Joint Conference on Artificial Intelligence (IJCAI)}, pages 1325--1331.

\bibitem[{Zhang et~al.(2022)Zhang, Li, and Kashima}]{moepra}
Guoxi Zhang, Jiyi Li, and Hisashi Kashima. 2022.
\newblock \href {https://doi.org/10.1109/DSAA54385.2022.10032454} {Improving pairwise rank aggregation via querying for rank difference}.
\newblock In \emph{2022 IEEE 9th International Conference on Data Science and Advanced Analytics (DSAA)}, pages 1--9.

\bibitem[{Zhou et~al.(2012)Zhou, Platt, Basu, and Mao}]{Minimax_Entropy}
Dengyong Zhou, John~C. Platt, Sumit Basu, and Yi~Mao. 2012.
\newblock \href {https://dl.acm.org/doi/10.5555/2999325.2999380} {Learning from the wisdom of crowds by minimax entropy}.
\newblock In \emph{Proceedings of the 25th International Conference on Neural Information Processing Systems (NIPS)}, pages 2195--2203.

\bibitem[{Zhu et~al.(2023)Zhu, Zhang, ul~Haq, Hui, and Tyson}]{CrowdChatgptSocialComputing}
Yiming Zhu, Peixian Zhang, Ehsan ul~Haq, Pan Hui, and Gareth Tyson. 2023.
\newblock \href {https://doi.org/10.48550/ARXIV.2304.10145} {Can chatgpt reproduce human-generated labels? {A} study of social computing tasks}.
\newblock \emph{CoRR}, abs/2304.10145.

\bibitem[{Zuo et~al.(2020)Zuo, Li, Zhou, Li, and Mao}]{affectipartialrank}
Xingkun Zuo, Jiyi Li, Qili Zhou, Jianjun Li, and Xiaoyang Mao. 2020.
\newblock \href {https://doi.org/10.1145/3394171.3413744} {Affecti: A game for diverse, reliable, and efficient affective image annotation}.
\newblock In \emph{Proceedings of the 28th ACM International Conference on Multimedia (MM)}, pages 529--537.

\end{thebibliography}
\bibliographystyle{acl_natbib}

\clearpage

\appendix
\section{Appendix}
\label{sec:appendix}
\begin{table}[!t]
\small
\setlength\tabcolsep{3pt}
\caption{\label{tab:aggregatorperformance_withGemini} Performance of different aggregators, C.A., L.A. (Gemini (G)), and Model Aggregator on the raw crowd answers $\mathcal{A_{C.C.}}$. }
\begin{center}
\begin{tabular}{c|cccccc}
\hline
\multirow{2}{*}{Data} & \multirow{2}{*}{C.A.} & \multicolumn{1}{c}{L.A.} & \multicolumn{3}{c}{Model Aggregator} \\
&  & (G) & SMV & SMS & RASA \\ 
\hline 
\multicolumn{6}{c}{(a) GLEU}
\\
J1 & 0.2124 & \textbf{0.2633} & 0.1930 & 0.2489 & 0.2537 \\ 
T1 & 0.1874 & 0.2015 & 0.1740 & 0.2310 & \textbf{0.2376} \\ 
T2 & 0.1817 & 0.1751 & 0.1616 & 0.2189 & \textbf{0.2340} \\ 
\hline
\multicolumn{6}{c}{(b) METEOR}
\\
J1 & 0.4156 & \textbf{0.4926} & 0.3861 & 0.4666 & 0.4745 \\  
T1 & 0.3909 & 0.4218 & 0.3786 & 0.4548 & \textbf{0.4718} \\  
T2 & 0.3840 & 0.3833 & 0.3604 & 0.4426 & \textbf{0.4653} \\ 
\hline
\multicolumn{6}{c}{(c) Embedding Similarity}
\\
J1 & 0.6857 & 0.7310 & 0.6732 & \textbf{0.7426} & 0.7414 \\ 
T1 & 0.7336 & 0.7635 & 0.7245 & 0.7935 & \textbf{0.8020} \\  
T2 & 0.7215 & 0.7540 & 0.7105 & 0.7860 & \textbf{0.7935} \\ 
\hline
\end{tabular}
\end{center}
\end{table}

\begin{table}[!t]
\scriptsize
\caption{\label{tab:aggresults_gleu_withGemini} Results of text answer aggregation: GLEU; LLM aggregator (Gemini (G)).}
\begin{center}
\begin{tabular}{cccc|c|cc}
\hline
\multicolumn{4}{c|}{Answers} & \multicolumn{3}{c}{Model Aggregator} \\
 & $\mathcal{A}_{C.C.}$ & $\mathcal{A}_{C.A.}$ & $\mathcal{A}_{L.A.}$ & SMV & SMS & RASA \\ 
\hline
\multicolumn{7}{c}{J1}
\\
\multirow{1}{*}{\Rmnum{1}} & $\bigcirc$ & & & 0.1930 & 0.2489 & 0.2537 \\
\cdashline{1-7}
\multirow{2}{*}{\Rmnum{2}} & & $\bigcirc$ & & 0.2260 & 0.2740 & 0.2444 \\
  & & & $\bigcirc$(G) & \textbf{0.2661} & 0.2623 & 0.2626 \\
\cdashline{1-7}
\multirow{1}{*}{\Rmnum{3}} & $\bigcirc$ & $\bigcirc$ & & 0.2040 & 0.2873 & 0.2708 \\
\cdashline{1-7}
\multirow{3}{*}{\Rmnum{4}} & $\bigcirc$ & & $\bigcirc$(G) & 0.2174 & 0.2856 & 0.2620 \\
 & & $\bigcirc$ & $\bigcirc$(G) & 0.2460 & 0.2746 & 0.2564 \\
 & $\bigcirc$ & $\bigcirc$ & $\bigcirc$(G) & 0.2195 & \textbf{0.2920} & \textbf{0.2723} \\
\hline
\multicolumn{7}{c}{T1}
\\
\multirow{1}{*}{\Rmnum{1}} & $\bigcirc$ & & & 0.1740 & 0.2310 & \textbf{0.2376} \\
\cdashline{1-7}
\multirow{2}{*}{\Rmnum{2}} & & $\bigcirc$ & & 0.1819 & 0.2074 & 0.2007 \\
 & & & $\bigcirc$(G) & \textbf{0.2082} & 0.2136 & 0.2126 \\
\cdashline{1-7}
\multirow{1}{*}{\Rmnum{3}} & $\bigcirc$ & $\bigcirc$ & & 0.1767 & 0.2257 & 0.2307 \\
\cdashline{1-7}
\multirow{3}{*}{\Rmnum{4}} & $\bigcirc$ & & $\bigcirc$(G) & 0.1854 & \textbf{0.2337} & 0.2138 \\
 & & $\bigcirc$ & $\bigcirc$(G) & 0.1950 & 0.2204 & 0.2169 \\
 & $\bigcirc$ & $\bigcirc$ & $\bigcirc$(G) & 0.1845 & 0.2274 & 0.2274 \\ 
\hline
\multicolumn{7}{c}{T2}
\\
\multirow{1}{*}{\Rmnum{1}} & $\bigcirc$ & & & 0.1616 & 0.2189 & 0.2340 \\
\cdashline{1-7}
\multirow{2}{*}{\Rmnum{2}} & & $\bigcirc$ & & 0.1769 & 0.2087 & 0.1952 \\
 & & & $\bigcirc$(G) & \textbf{0.1837} & 0.1853 & 0.1707 \\
\cdashline{1-7}
\multirow{1}{*}{\Rmnum{3}} & $\bigcirc$ & $\bigcirc$ & & 0.1667 & \textbf{0.2369} & \textbf{0.2341} \\
\cdashline{1-7}
\multirow{3}{*}{\Rmnum{4}} & $\bigcirc$ & & $\bigcirc$(G) & 0.1690 & 0.2174 & 0.2032 \\
 & & $\bigcirc$ & $\bigcirc$(G) & 0.1803 & 0.2061 & 0.2191 \\
 & $\bigcirc$ & $\bigcirc$ & $\bigcirc$(G) & 0.1710 & 0.2202 & 0.2277 \\
\hline
\end{tabular}\\
\end{center}
\end{table}

\begin{table}[!t]
\scriptsize
\caption{\label{tab:aggresults_meteor_withGemini} Results of text answer aggregation: METEOR; LLM aggregator (Gemini (G)).}
\begin{center}
\begin{tabular}{cccc|c|cc}
\hline
\multicolumn{4}{c|}{Answers} & \multicolumn{3}{c}{Model Aggregator} \\
 & $\mathcal{A}_{C.C.}$ & $\mathcal{A}_{C.A.}$ & $\mathcal{A}_{L.A.}$ & SMV & SMS & RASA \\ 
\hline
\multicolumn{7}{c}{J1}
\\
\multirow{1}{*}{\Rmnum{1}} & $\bigcirc$ & & & 0.3861 & 0.4666 & 0.4745 \\ \cdashline{1-7}
\multirow{2}{*}{\Rmnum{2}} & & $\bigcirc$ & & 0.4397 & 0.4912 & 0.4524 \\
 & & & $\bigcirc$(G) & \textbf{0.4954} & 0.4925 & 0.4885 \\
\cdashline{1-7}
\multirow{1}{*}{\Rmnum{3}} & $\bigcirc$ & $\bigcirc$ & & 0.4040 & 0.5092 & 0.4946 \\
\cdashline{1-7}
\multirow{3}{*}{\Rmnum{4}} & $\bigcirc$ & & $\bigcirc$(G) & 0.4226 & 0.5142 & 0.4911 \\
 & & $\bigcirc$ & $\bigcirc$(G) & 0.4675 & 0.5084 & 0.4878 \\
 & $\bigcirc$ & $\bigcirc$ & $\bigcirc$(G) & 0.4269 & \textbf{0.5204} & \textbf{0.5037} \\
\hline
\multicolumn{7}{c}{T1}
\\
\multirow{1}{*}{\Rmnum{1}} & $\bigcirc$ & & & 0.3786 & 0.4548 & \textbf{0.4718} \\ \cdashline{1-7}
\multirow{2}{*}{\Rmnum{2}} & & $\bigcirc$ & & 0.3927 & 0.4330 & 0.4272 \\
 & & & $\bigcirc$(G) & \textbf{0.4258} & 0.4364 & 0.4392 \\
\cdashline{1-7}
\multirow{1}{*}{\Rmnum{3}} & $\bigcirc$ & $\bigcirc$ & & 0.3833 & 0.4528 & 0.4564 \\
\cdashline{1-7}
\multirow{3}{*}{\Rmnum{4}} & $\bigcirc$ & & $\bigcirc$(G) & 0.3943 & \textbf{0.4630} & 0.4447 \\
 & & $\bigcirc$ & $\bigcirc$(G) & 0.4092 & 0.4395 & 0.4379 \\
 & $\bigcirc$ & $\bigcirc$ & $\bigcirc$(G) & 0.3939 & 0.4584 & 0.4514 \\ 
\hline
\multicolumn{7}{c}{T2}
\\
\multirow{1}{*}{\Rmnum{1}} & $\bigcirc$ & & & 0.3604 & 0.4426 & \textbf{0.4653} \\ \cdashline{1-7}
\multirow{2}{*}{\Rmnum{2}} & & $\bigcirc$ & & 0.3746 & 0.4238 & 0.4083 \\
 & & & $\bigcirc$(G) & \textbf{0.3956} & 0.3999 & 0.3754 \\
\cdashline{1-7}
\multirow{1}{*}{\Rmnum{3}} & $\bigcirc$ & $\bigcirc$ & & 0.3651 & \textbf{0.4507} & 0.4551 \\
\cdashline{1-7}
\multirow{3}{*}{\Rmnum{4}} & $\bigcirc$ & & $\bigcirc$(G) & 0.3721 & 0.4426 & 0.4179 \\
 & & $\bigcirc$ & $\bigcirc$(G) & 0.3851 & 0.4200 & 0.4324 \\
 & $\bigcirc$ & $\bigcirc$ & $\bigcirc$(G) & 0.3727 & 0.4384 & 0.4469 \\
\hline
\end{tabular}\\
\end{center}
\end{table}

\begin{table}[t!]
\scriptsize
\caption{\label{tab:aggresults_embedding_withGemini} Results of text answer aggregation: Embedding Similarity; LLM aggregator (Gemini (G)).}
\begin{center}
\begin{tabular}{cccc|c|cc}
\hline
\multicolumn{4}{c|}{Answers} & \multicolumn{3}{c}{Model Aggregator} \\
 & $\mathcal{A}_{C.C.}$ & $\mathcal{A}_{C.A.}$ & $\mathcal{A}_{L.A.}$ & SMV & SMS & RASA \\ 
\hline
\multicolumn{7}{c}{J1}
\\
\multirow{1}{*}{\Rmnum{1}} & $\bigcirc$ & & & 0.6732 & 0.7426 & 0.7414 \\
\cdashline{1-7}
\multirow{2}{*}{\Rmnum{2}} & & $\bigcirc$ & & 0.6983 & 0.7360 & 0.7188 \\
 & & & $\bigcirc$(G) & \textbf{0.7310} & 0.7313 & 0.7311 \\
\cdashline{1-7}
\multirow{1}{*}{\Rmnum{3}} & $\bigcirc$ & $\bigcirc$ & & 0.6815 & 0.7516 & \textbf{0.7459} \\
\cdashline{1-7}
\multirow{3}{*}{\Rmnum{4}} & $\bigcirc$ & & $\bigcirc$(G) & 0.6925 & 0.7463 & 0.7311 \\
 & & $\bigcirc$ & $\bigcirc$(G) & 0.7147 & 0.7391 & 0.7315 \\
 & $\bigcirc$ & $\bigcirc$ & $\bigcirc$(G) & 0.6939 & \textbf{0.7521} & 0.7423 \\
\hline
\multicolumn{7}{c}{T1}
\\
\multirow{1}{*}{\Rmnum{1}} & $\bigcirc$ & & & 0.7245 & 0.7935 & \textbf{0.8020} \\
\cdashline{1-7}
\multirow{2}{*}{\Rmnum{2}} & & $\bigcirc$ & & 0.7430 & 0.7851 & 0.7758 \\
\cdashline{4-4}
 & & & $\bigcirc$(G) & \textbf{0.7635} & 0.7698 & 0.7701 \\
\cdashline{1-7}
\multirow{1}{*}{\Rmnum{3}} & $\bigcirc$ & $\bigcirc$ & & 0.7306 & 0.7965 & 0.7976 \\
\cdashline{1-7}
\multirow{3}{*}{\Rmnum{4}}  & $\bigcirc$ & & $\bigcirc$(G) & 0.7375 & \textbf{0.7978} & 0.7762 \\
 & & $\bigcirc$ & $\bigcirc$(G) & 0.7532 & 0.7803 & 0.7803 \\
 & $\bigcirc$ & $\bigcirc$ & $\bigcirc$(G) & 0.7389 & 0.7963 & 0.7953 \\ 
\hline
\multicolumn{7}{c}{T2}
\\
\multirow{1}{*}{\Rmnum{1}} & $\bigcirc$ & & & 0.7105 & 0.7860 & \textbf{0.7935} \\ \cdashline{1-7}
\multirow{2}{*}{\Rmnum{2}} &  & $\bigcirc$ & & 0.7282 & 0.7681 & 0.7468 \\
 & & & $\bigcirc$(G) & \textbf{0.7540} & 0.7579 & 0.7468 \\
\cdashline{1-7}
\multirow{1}{*}{\Rmnum{3}} & $\bigcirc$ & $\bigcirc$ & & 0.7164 & \textbf{0.7885} & 0.7822 \\
\cdashline{1-7}
\multirow{3}{*}{\Rmnum{4}} & $\bigcirc$ & & $\bigcirc$(G) & 0.7250 & 0.7856 & 0.7744 \\
 & & $\bigcirc$ & $\bigcirc$(G) & 0.7411 & 0.7679 & 0.7724 \\
 & $\bigcirc$ & $\bigcirc$ & $\bigcirc$(G) & 0.7258 & 0.7833 & 0.7831 \\
\hline
\end{tabular}\\
\end{center}
\end{table}

\begin{table*}[!t]
\small
\setlength\tabcolsep{1.5pt}
\caption{\label{tab:singleworkerresults_withGemini} Quality of individual answers by Crowd Creators (C.C.), Crowd Aggregators (C.A.) and LLM Aggregators (L.A., Gemini (G)) on GLEU, METEOR and embedding similarity. }
\begin{center}
(a) GLEU\\
\begin{tabular}{c|ccc|ccc|ccc|ccc|ccc}
\hline
\multirow{3}{*}{Data} & \multicolumn{3}{c|}{MIN} & \multicolumn{3}{c|}{MEAN} & \multicolumn{3}{c|}{MAX} & \multicolumn{3}{c|}{STD} & \multicolumn{3}{c}{TIAA} \\ 
 & C.C. & C.A. & L.A.(G) & C.C. & C.A. & L.A.(G) & C.C. & C.A. & L.A.(G) & C.C. & C.A. & L.A.(G) & C.C. & C.A. & L.A.(G) \\ 
\hline
J1 & 0.0724 & 0.0000 & \textbf{0.2588} & 0.1868 & 0.2124 & \textbf{0.2633} & 0.5948 & \textbf{1.0000} & 0.2714 & 0.0915 & 0.1399 & 0.0057 & 0.1798 & 0.2746 & 0.7749 \\
T1 & 0.0669 & 0.0217 & \textbf{0.1968} & 0.1764 & 0.1874 & \textbf{0.2015} & 0.5534 & \textbf{0.7895} & 0.2064 & 0.0818 & 0.1297 & 0.0039 & 0.2533 & 0.3473 & 0.8028 \\
T2 & 0.0503 & 0.0000 & \textbf{0.1707} & 0.1716 & \textbf{0.1817} & 0.1751 & 0.4540 & \textbf{0.6842} & 0.1810 & 0.0838 & 0.1190 & 0.0043 & 0.2377 & 0.3371 & 0.7854 \\
\hline
\end{tabular}
\vspace{0.1cm}\\
(b) METEOR\\
\begin{tabular}{c|ccc|ccc|ccc|ccc|ccc}
\hline
\multirow{2}{*}{Data} & \multicolumn{3}{c|}{MIN} & \multicolumn{3}{c|}{MEAN} & \multicolumn{3}{c|}{MAX} & \multicolumn{3}{c|}{STD} & \multicolumn{3}{c}{TIAA}\\ 
 & C.C. & C.A. & L.A.(G) & C.C. & C.A. & L.A.(G) & C.C. & C.A. & L.A.(G) & C.C. & C.A. & L.A.(G) & C.C. & C.A. & L.A.(G)\\ 
\hline
J1 & 0.2041 & 0.0490 & \textbf{0.4869} & 0.3762 & 0.4156 & \textbf{0.4926} & 0.7464 & \textbf{0.9985} & 0.4993 & 0.1037 & 0.1569 & 0.0051 & 0.3791 & 0.4910 & 0.8792 \\
T1 & 0.1781 & 0.0394 & \textbf{0.4165} & 0.3771 & 0.3909 & \textbf{0.4218} & 0.7163 & \textbf{0.9170} & 0.4310 & 0.0987 & 0.1666 & 0.0065 & 0.4763 & 0.5670 & 0.8830 \\
T2 & 0.1523 & 0.0476 & \textbf{0.3754} & 0.3630 & \textbf{0.3840} & 0.3833 & 0.6473 & \textbf{0.8288} & 0.3941 & 0.1010 & 0.1619 & 0.0078 & 0.4402 & 0.5492 & 0.8728 \\
\hline
\end{tabular}
\vspace{0.1cm}\\
(c) Embedding Similarity\\
\begin{tabular}{c|ccc|ccc|ccc|ccc|ccc}
\hline
\multirow{2}{*}{Data} & \multicolumn{3}{c|}{MIN} & \multicolumn{3}{c|}{MEAN} & \multicolumn{3}{c|}{MAX} & \multicolumn{3}{c|}{STD} & \multicolumn{3}{c}{TIAA} \\ 
 & C.C. & C.A. & L.A.(G) & C.C. & C.A. & L.A.(G) & C.C. & C.A. & L.A.(G) & C.C. & C.A. & L.A.(G) & C.C. & C.A. & L.A.(G) \\ 
\hline
J1 & 0.4233 & 0.2243 & \textbf{0.7265} & 0.6620 & 0.6857 & \textbf{0.7310} & 0.8795 & \textbf{1.0000} & 0.7373 & 0.0687 & 0.1135 & 0.0036 & 0.6583 & 0.7286 & 0.9373 \\
T1 & 0.5827 & 0.1282 & \textbf{0.7566} & 0.7260 & 0.7336 & \textbf{0.7635} & 0.9043 & \textbf{1.0000} & 0.7719 & 0.0687 & 0.1511 & 0.0058 & 0.7252 & 0.7850 & 0.9502 \\
T2 & 0.4726 & 0.2700 & \textbf{0.7458} & 0.7077 & 0.7215 & \textbf{0.7540} & 0.8958 & \textbf{1.0000} & 0.7684 & 0.0861 & 0.1250 & 0.0086 & 0.7060 & 0.7752 & 0.9360 \\
\hline
\end{tabular}
\end{center}
\end{table*}

\begin{figure*}[!t]
\begin{minipage}{0.315\textwidth}
\centering
\includegraphics[height=3.5cm]{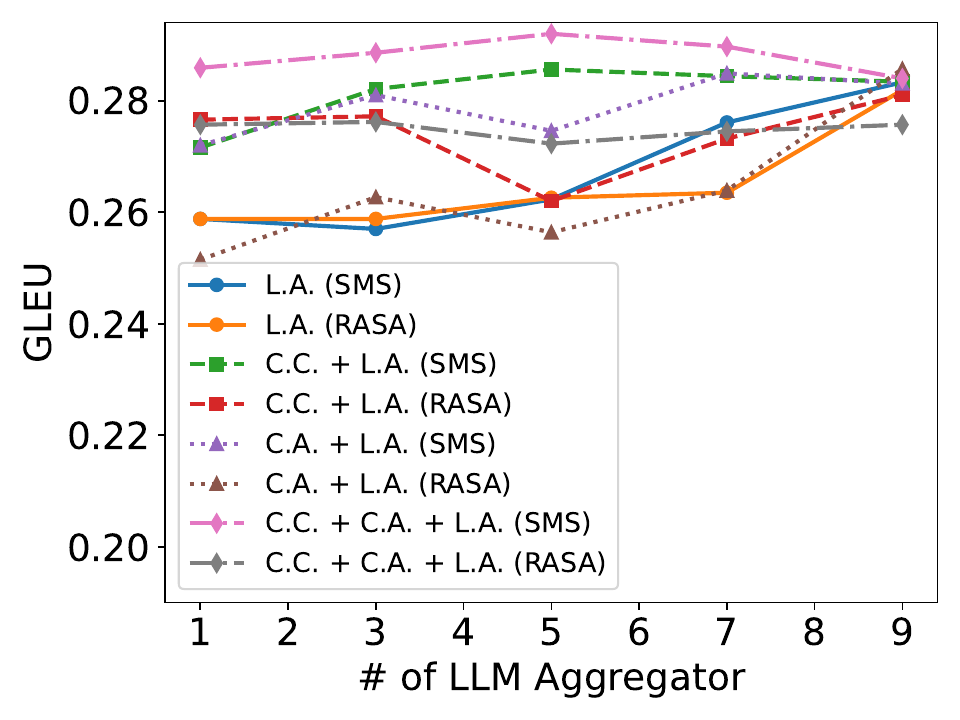}
\\(a) J1
\end{minipage}
\begin{minipage}{0.315\textwidth}
\centering
\includegraphics[height=3.5cm]{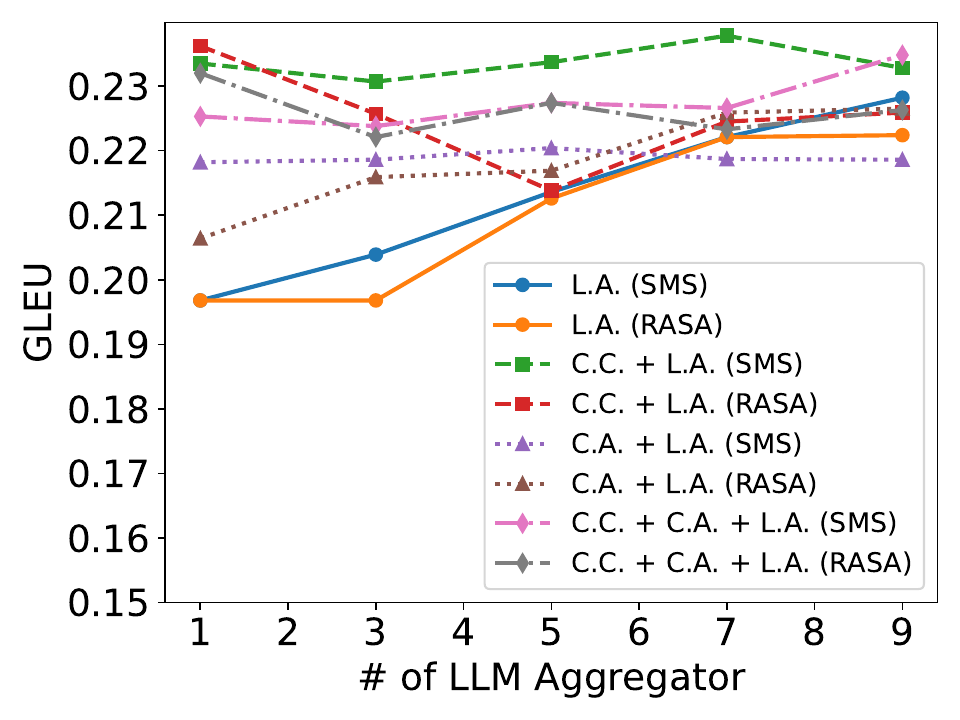}
\\(b) T1
\end{minipage}
\begin{minipage}{0.315\textwidth}
\centering
\includegraphics[height=3.5cm]{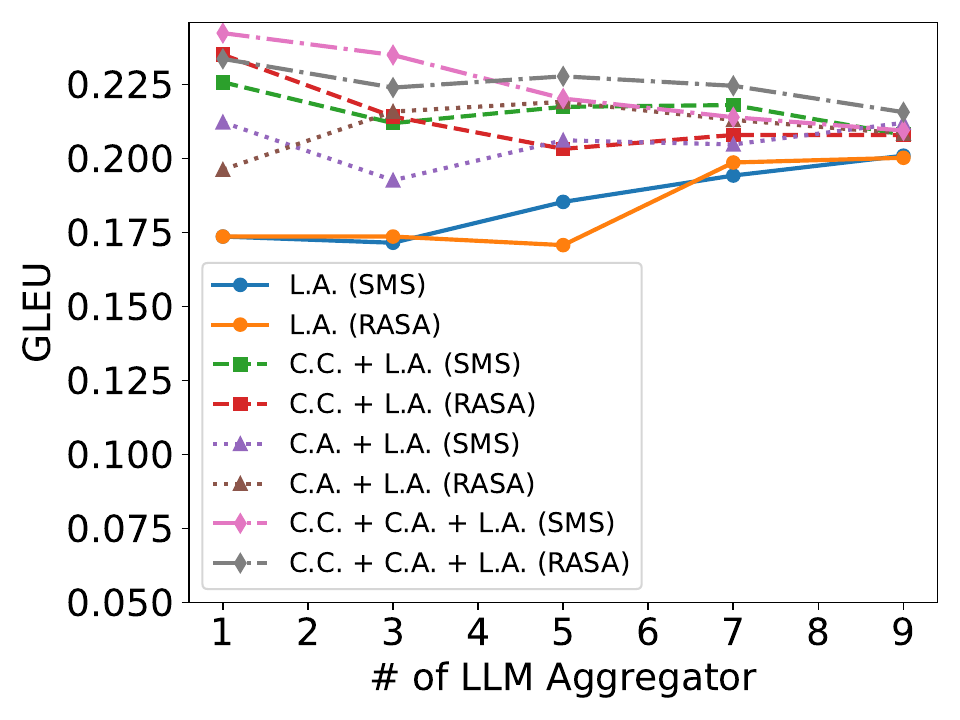}
\\(c) T2
\end{minipage}
\caption{GLEU Results by different number of L.A. (Gemini) based on SMS and RASA. }
\label{fig:diff_num_llmagg_gleu_gemini} 
\end{figure*}

\begin{figure*}[!t]
\begin{minipage}{0.315\textwidth}
\centering
\includegraphics[height=3.5cm]{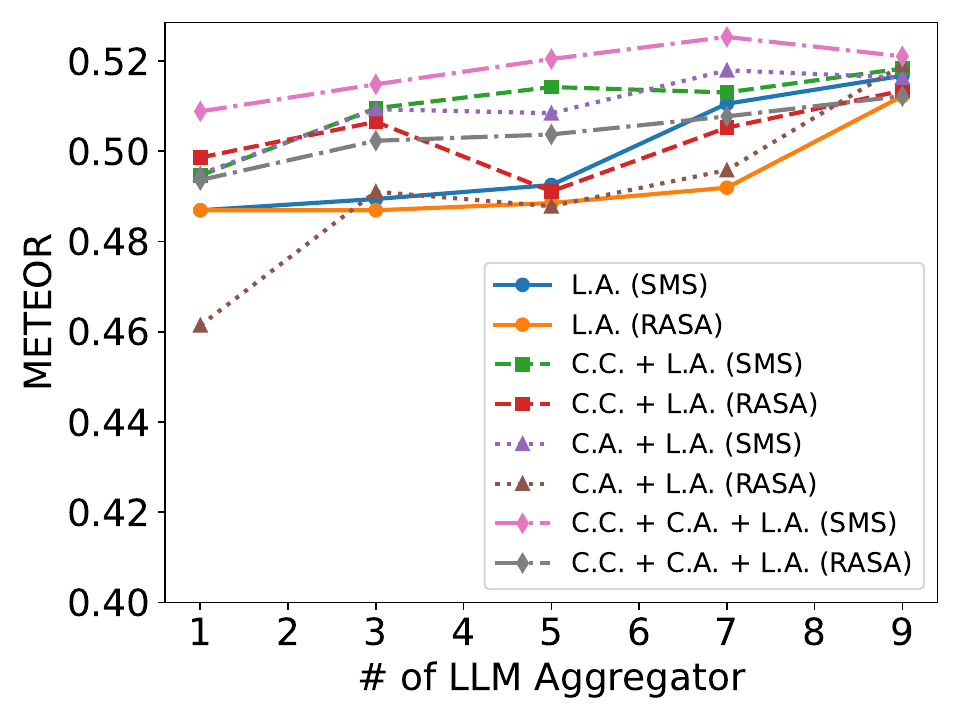}
\\(a) J1
\end{minipage}
\begin{minipage}{0.315\textwidth}
\centering
\includegraphics[height=3.5cm]{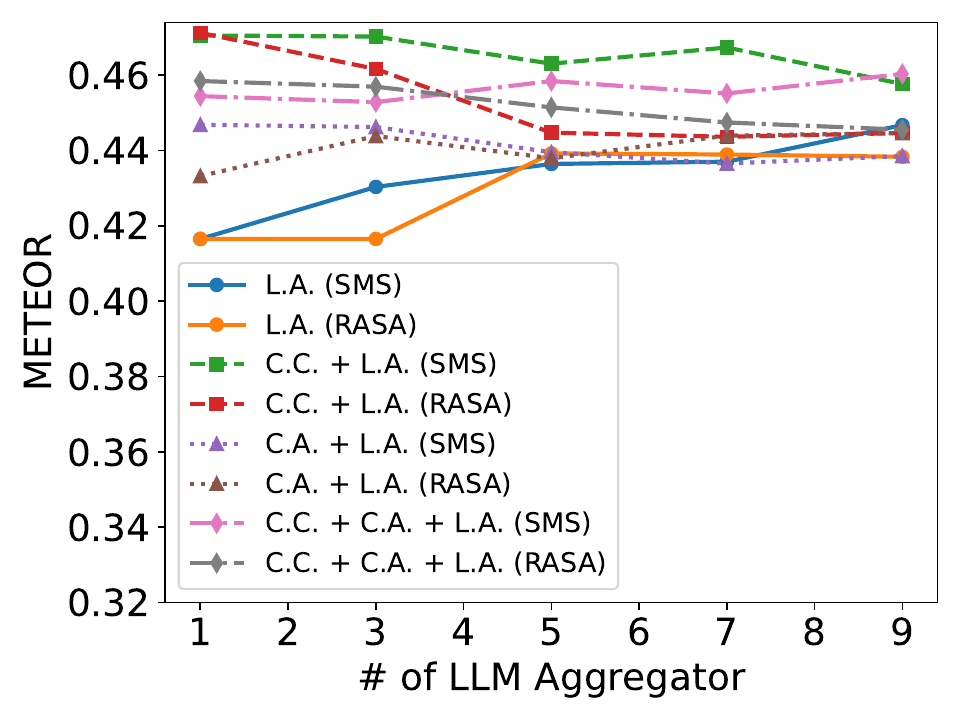}
\\(b) T1
\end{minipage}
\begin{minipage}{0.315\textwidth}
\centering
\includegraphics[height=3.5cm]{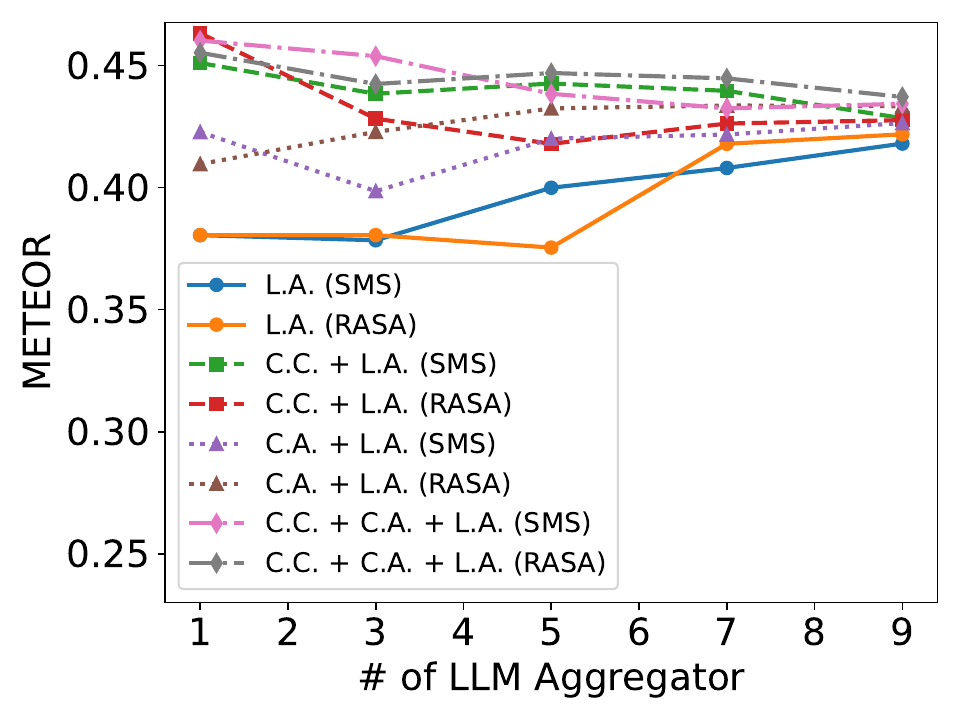}
\\(c) T2
\end{minipage}
\caption{METEOR Results by different number of L.A. (Gemini) based on SMS and RASA. }
\label{fig:diff_num_llmagg_meteor_gemini} 
\end{figure*}

\begin{figure*}[!t]
\begin{minipage}{0.315\textwidth}
\centering
\includegraphics[height=3.5cm]{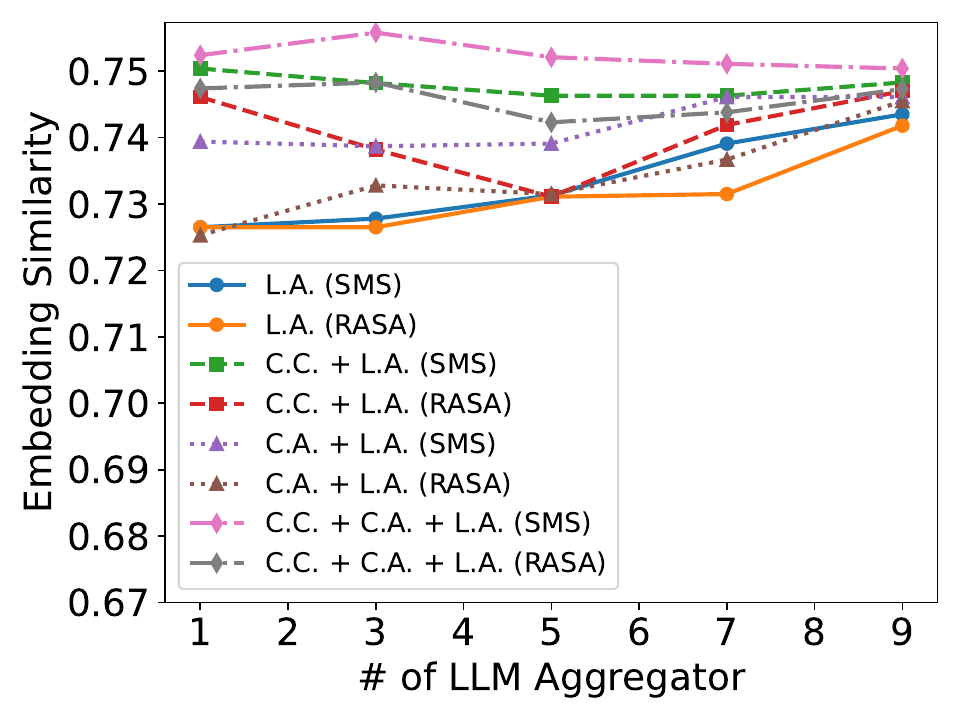}
\\(a) J1
\end{minipage}
\begin{minipage}{0.315\textwidth}
\centering
\includegraphics[height=3.5cm]{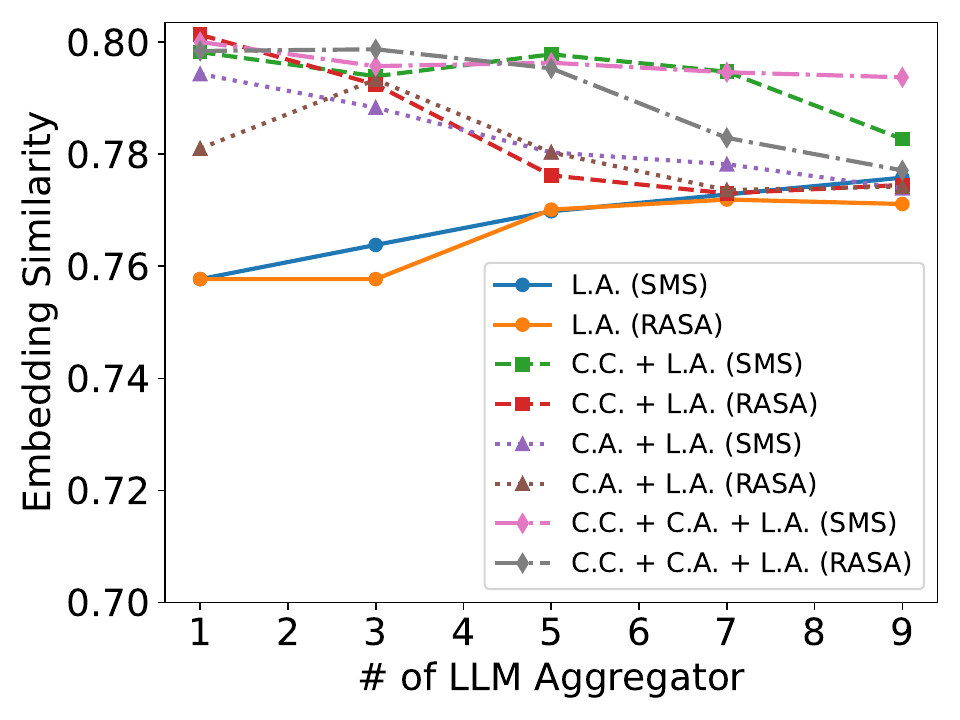}
\\(b) T1
\end{minipage}
\begin{minipage}{0.315\textwidth}
\centering
\includegraphics[height=3.5cm]{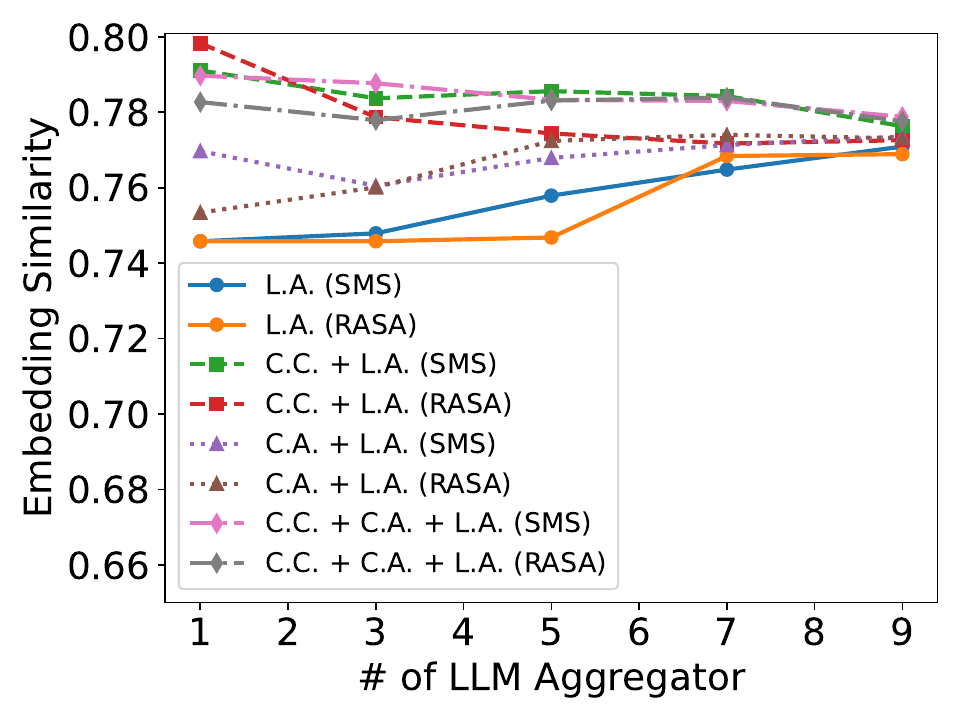}
\\(c) T2
\end{minipage}
\caption{Embedding Similarity Results by different number of L.A. (Gemini) based on SMS and RASA. }
\label{fig:diff_num_llmagg_embedsim_gemini} 
\end{figure*}

\section{Additional Results By Gemini}
\textit{Because the Gemini results are almost always worse than GPT-4 results, we only put the GPT-4 results in the main manuscript and put the Gemini results in the Appendix Section. } 

Table \ref{tab:singleworkerresults_withGemini} lists the results of the individual performance of crowd creators, crowd aggregators, and Gemini LLM aggregators from the viewpoint of the quality of answers they provide or generate. 

We also make a comparison among different types of aggregators on the raw crowd answers. We use the mean results of crowd aggregators and LLM aggregators in Table \ref{tab:singleworkerresults_withGemini} and add the results of the model aggregator to list them in Table \ref{tab:aggregatorperformance_withGemini}. 

Table \ref{tab:aggresults_gleu_withGemini} shows the results on text answer aggregation on the evaluation metric GLEU. $\mathcal{A}_{L.A.}$ with (G) uses Gemini.  
Table \ref{tab:aggresults_meteor_withGemini} and \ref{tab:aggresults_embedding_withGemini} list the results on the evaluation metrics METEOR and Embedding Similarity. They reach similar observations with the evaluation metric GLEU.  

Figure \ref{fig:diff_num_llmagg_gleu_gemini}, \ref{fig:diff_num_llmagg_meteor_gemini} and \ref{fig:diff_num_llmagg_embedsim_gemini} shows the results on GLEU, METEOR and Embedding Similarity by different numbers of Gemini LLM aggregators based on model aggregators SMS and RASA. Temperature of 1 LLM aggregator: [0]; temperatures of 3 LLM aggregators: [0, 0.5, 1]; temperatures of 5 LLM aggregators: [0, 0.25, 0.5, 0.75, 1]; temperatures of 7 LLM aggregators: [0, 0.1, 0.25, 0.5, 0.75, 0.9, 1]; temperatures of 9 LLM aggregators: [0, 0.1, 0.25, 0.5, 0.75, 0.9, 1, 1.1, 1.25]. 
They reach similar observations with the evaluations GPT-4 LLM aggregators in the main manuscript.

\end{document}